\documentclass{article} 
\usepackage{iclr2027_conference,times}

\usepackage{amsmath,amsfonts,bm}

\def\eqref#1{equation~\ref{#1}}

\def\1{\bm{1}}

\DeclareMathAlphabet{\mathsfit}{\encodingdefault}{\sfdefault}{m}{sl}
\SetMathAlphabet{\mathsfit}{bold}{\encodingdefault}{\sfdefault}{bx}{n}

\usepackage{hyperref}
\usepackage{url}
\usepackage{xcolor}
\usepackage[table]{xcolor}

\usepackage{graphicx}
\usepackage{booktabs}
\usepackage{amssymb}
\usepackage{pifont}
\newcommand{\xmark}{\ding{55}}
\usepackage{wrapfig}
\usepackage{multirow}
\usepackage{caption}
\usepackage{algorithm}
\usepackage{algpseudocode}
\usepackage{enumitem}

\title{HARMONI: Aligning Human and Scene Priors \\for Multi-View 4D Reconstruction}

\author{Sangmin Kim\textsuperscript{1}\thanks{This work was done during an internship at NAVER Cloud.} \quad Minhyuk Hwang\textsuperscript{1} \quad Geonho Cha\textsuperscript{2} \quad Dongyoon Wee\textsuperscript{2} \quad Jaesik Park\textsuperscript{1$\dagger$} \\
\textsuperscript{1}Seoul National University \quad \textsuperscript{2}NAVER Cloud
}

\iclrfinalcopy 
\begin{document}

\maketitle
\newcommand{\ours}{HARMONI}
\begingroup
\renewcommand\thefootnote{$\dagger$}
\footnotetext{Corresponding author.}
\endgroup
\begin{figure}[!h]
  \centering
  \includegraphics[width=0.95\linewidth]{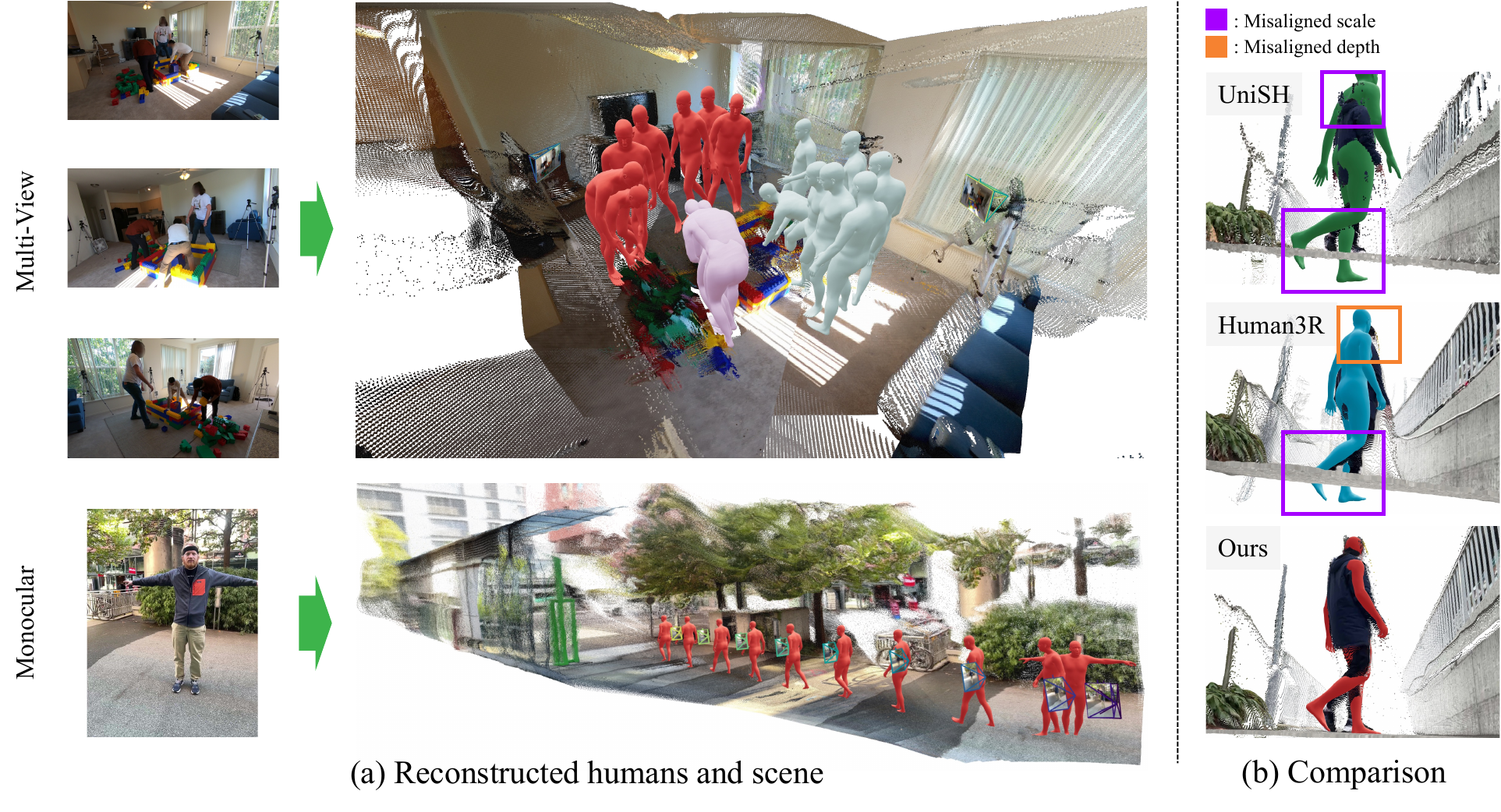}
  \caption{(a) We present \textbf{\ours}, which reconstructs cameras, scene, and humans with identities from multi-view or monocular videos without test-time optimization. 
  (b) With \emph{bidirectional anchoring}, our model outperforms previous human-scene reconstruction methods in both settings.}
  \label{fig:teaser}
\end{figure}
\begin{abstract}
Recent advances in 3D foundation models have enabled joint reconstruction of humans and their surrounding environments.
However, combining independently trained human and scene priors often produces misalignment in scale and depth.
We observe that the two priors have complementary strengths. The scene prior provides consistent depth but approximate scale, while the human prior provides fixed body scale but less reliable depth.
Motivated by this observation, we present \textbf{\ours}, a feed-forward framework that reconstructs cameras, scene, and humans with identities from monocular or multi-view video without test-time optimization.
We introduce \emph{bidirectional anchoring}, in which scene depth guides human placement, while human keypoints calibrate the scene scale to match the human body.
While most existing approaches target monocular inputs and multi-view methods rely on optimization or re-identification, our approach naturally extends to multiple views.
Building on bidirectional anchoring, we introduce a \emph{multi-view fusion module} that merges per-view estimates, while reducing the influence of unreliable views and visually similar individuals.
Experiments show that our framework outperforms previous human-scene methods in global motion and multi-view pose estimation by up to 28\% and 64\%, while running 28$\times$ faster than optimization-based approaches.
\end{abstract}
\section{Introduction}
\label{sec:intro}
Understanding how people interact with 3D environments requires reconstructing both humans and their surroundings in a shared coordinate system. Joint reconstruction of humans and scene enables downstream applications such as training humanoids~\citep{videomimic, wang2026contact}, simulating scenes for autonomous driving~\citep{chen2025omnire}, and creating AR/VR entertainment~\citep{kim2025showmak3r, hwang2026studiorecon}.
With advances in 3D foundation models such as DUSt3R~\citep{wang2024dust3r} and VGGT~\citep{wang2025vggt}, many works have tried to jointly recover humans and their environments by combining 3D foundation models with human mesh recovery models~\citep{baradel2024multi}. Recent approaches such as UniSH~\citep{li2026unish} and Human3R~\citep{chen2025human3r} unify human and scene reconstruction into a single feed-forward architecture. 

However, one of the challenges in jointly reconstructing humans and scene is that the two can be misaligned. This issue arises in two aspects: scale and depth.
Since 3D foundation models \textit{estimate} the metric scale of the scene, it can differ from the scale of parametric human meshes such as SMPL~\citep{SMPL:2015}, which are \textit{defined} in metric units. 
In addition, because the scene and human priors are trained independently, the depth of the scene points corresponding to the human can differ from the depth given by the \(z\)-component of the SMPL global translation.
If the scale is inconsistent, reconstructed humans can float above or penetrate the ground, and if the depth is inconsistent, they can be placed too close to or too far from the camera relative to the scene (Fig.~\ref{fig:teaser}b).

Moreover, most existing works target monocular inputs, and extending to multi-view introduces additional challenges. Predictions from different views must be expressed in a shared coordinate system, with humans and scene aligned within each view and person correspondences established across views. Existing multi-view approaches address these requirements through additional stages. HSfM~\citep{muller2025reconstructing} relies on external 2D keypoint estimation, known cross-view person identities, and iterative optimization, whereas HAMSt3R~\citep{rojas2025hamst3r} obtains SMPL meshes through a fitting procedure. These stages increase system complexity and computational cost.

To address these issues, we present \textbf{\ours}~(aligning \textbf{H}uman \textbf{A}nd scene p\textbf{R}iors for \textbf{M}ulti-view 4D Rec\textbf{ON}struct\textbf{I}on), a feed-forward framework that reconstructs cameras, the scene, and humans with identities from monocular or multi-view videos without test-time optimization (Fig.~\ref{fig:teaser}a). We observe that human and scene priors have complementary strengths. The scene prior provides consistent depth but only approximate scale, whereas the human prior provides a fixed body scale but less reliable depth. 
Motivated by this observation, we introduce \emph{bidirectional anchoring} that exploits these complementary strengths. Human keypoints calibrate the scene scale to the human body, while scene depth guides global human placement. 
This alignment places humans coherently within the scene and improves both global motion and multi-view pose accuracy (Fig.~\ref{fig:teaser}b).

Since bidirectional anchoring aligns human and scene predictions within each view, our framework can be naturally extended to multi-view inputs. We introduce a \emph{multi-view fusion module} that aggregates per-view human estimates into a global representation. Specifically, confidence-weighted averaging suppresses unreliable per-view estimates, triangulation recovers precise global positions, and a refine transformer refines the fused estimates. Our module establishes cross-view person correspondences through geometry-based association using 3D positions and human poses. Unlike appearance-based re-identification~\citep{li2024multi}, our association remains robust for visually similar individuals. Together, our module enables optimization-free multi-view multi-person fusion.

Experiments on a monocular~\citep{kaufmann2023emdb} dataset and multi-view~\citep{huang2022capturing, khirodkar2023ego, grauman2024ego} datasets demonstrate that our framework outperforms previous human–scene reconstruction methods, reducing global motion error by up to 28\% and multi-view pose error by up to 64\%, while running 28$\times$ faster than optimization-based baselines.

Our main contributions are summarized as follows:
\begin{itemize}
\item We present \textbf{\ours}, a feed-forward framework that recovers cameras, scenes, and humans with identities from monocular and multi-view video without test-time optimization.
\item We propose \emph{bidirectional anchoring}, which aligns human and scene priors: human keypoints calibrate scene scale, while scene depth guides human placement.
\item We introduce a \emph{multi-view fusion module} that merges per-view estimates without test-time optimization, while reducing the influence of unreliable views.
\item Our framework reduces global motion and multi-view pose errors by up to 28\% and 64\%, respectively, and runs over 28$\times$ faster than optimization-based approaches.
\end{itemize}
\section{Related Work}
\vspace{2mm}\noindent\textbf{3D Scene Reconstruction.} Classical 3D reconstruction approaches such as Structure-from-Motion (SfM)~\citep{longuet1981computer, tomasi1992shape, beardsley19963d, fitzgibbon1998automatic, schonberger2016structure} and Multi-View Stereo (MVS)~\citep{seitz2006comparison, furukawa2009accurate, schonberger2016pixelwise} estimate camera poses and scene geometry through feature matching and bundle adjustment. While highly accurate, these optimization-based pipelines are computationally expensive and struggle in dynamic scenes.
Recent data-driven approaches replace iterative optimization with feed-forward neural architectures. DUSt3R~\citep{wang2024dust3r} predicts a 3D point map from an image pair in the first camera's coordinate system, enabling efficient reconstruction, and MASt3R~\citep{leroy2024grounding} improves matching stability and reconstruction accuracy.
VGGT~\citep{wang2025vggt} replaces explicit camera estimation with direct network prediction, while Pi3~\citep{wang2025pi} introduces a permutation-equivariant architecture to improve robustness. 
MapAnything~\citep{keetha2025mapanything} and Depth Anything 3~\citep{lin2025depth} extend feed-forward reconstruction toward metric-scale prediction.
More recently, VGGT-$\Omega$~\citep{wang2026vggt} further improves reconstruction accuracy and efficiency for both static and dynamic scenes.

\vspace{2mm}\noindent\textbf{Human Mesh Recovery.} 
Human Mesh Recovery (HMR) predicts parameters of body models such as SMPL~\citep{SMPL:2015} and SMPL-X~\citep{SMPL-X:2019} from images. HMR2.0~\citep{goel2023humans} employs a fully transformer-based architecture for single-image HMR, while Multi-HMR~\citep{baradel2024multi} enables single-shot multi-person whole-body reconstruction. Multi-HMR~2~\citep{fiche2026multi} further adopts a DETR-style decoder with camera intrinsics prediction.
While these methods operate in the camera coordinate frame, recent research has shifted toward reconstructing humans in the world coordinate system. 
SLAHMR~\citep{ye2023decoupling} separately reconstructs camera and human motion and then aligns them in a shared world coordinate system through optimization.
WHAM~\citep{shin2024wham} improves world-grounded motion estimation through motion priors and contact-aware refinement. 
GVHMR~\citep{shen2024world} introduces a gravity-aligned intermediate coordinate system to stabilize world-space motion prediction. 
TRAM~\citep{wang2024tram} combines SLAM-based camera estimation with transformer-based motion regression to reconstruct metric-scale human trajectories in world coordinates.
These approaches enable world-grounded human motion reconstruction from monocular video. 
However, they focus on human motion rather than jointly reconstructing humans and surrounding scene geometry.

\vspace{2mm}\noindent\textbf{Human-Scene Reconstruction.}
Recent work attempts to combine 3D foundation models with human mesh recovery to jointly reconstruct humans and their surroundings.
JOSH~\citep{liu2025joint} reconstructs humans and scenes from monocular videos using foot-contact cues for global alignment, and its extension JOSH3R further adopts a feed-forward architecture.
UniSH~\citep{li2026unish} integrates Pi3 and CameraHMR~\citep{patel2025camerahmr}, and finetunes the backbone and trains AlignNet, which requires large-scale internet data.
Human3R~\citep{chen2025human3r} targets unified human-scene reconstruction, but directly regresses human translation and relies on the scale estimate of CUT3R~\citep{wang2025continuous}.
MetricHMSR~\citep{song2026metrichmsr} and GRAFT~\citep{pradyumna2026graft} recover humans and scenes from a single image, but do not model cameras.
Since these approaches operate in monocular settings, several works explore multi-view reconstruction.
HSfM~\citep{muller2025reconstructing} uses human joints as cross-view correspondences to recover cameras and metric scale, but does so through iterative optimization with external 2D keypoints and requires known person identities. HAMSt3R~\citep{rojas2025hamst3r} predicts segmentation, dense pose, and scene geometry in a feed-forward manner at a single timestep, but requires additional fitting to obtain SMPL meshes. 
Trophies~\citep{liu2026trophies} handles multi-view video, but still relies on test-time optimization for cross-view alignment.
While HSfM and JOSH exploit humans for global alignment, they rely on test-time optimization. In contrast, our bidirectional anchoring calibrates the scene scale from body proportions and anchors human depth on the scene prior in a single feed-forward pass, providing aligned per-view estimates for our optimization-free multi-view 
fusion.


\begin{figure}[!t]
  \centering
  \includegraphics[width=\linewidth]{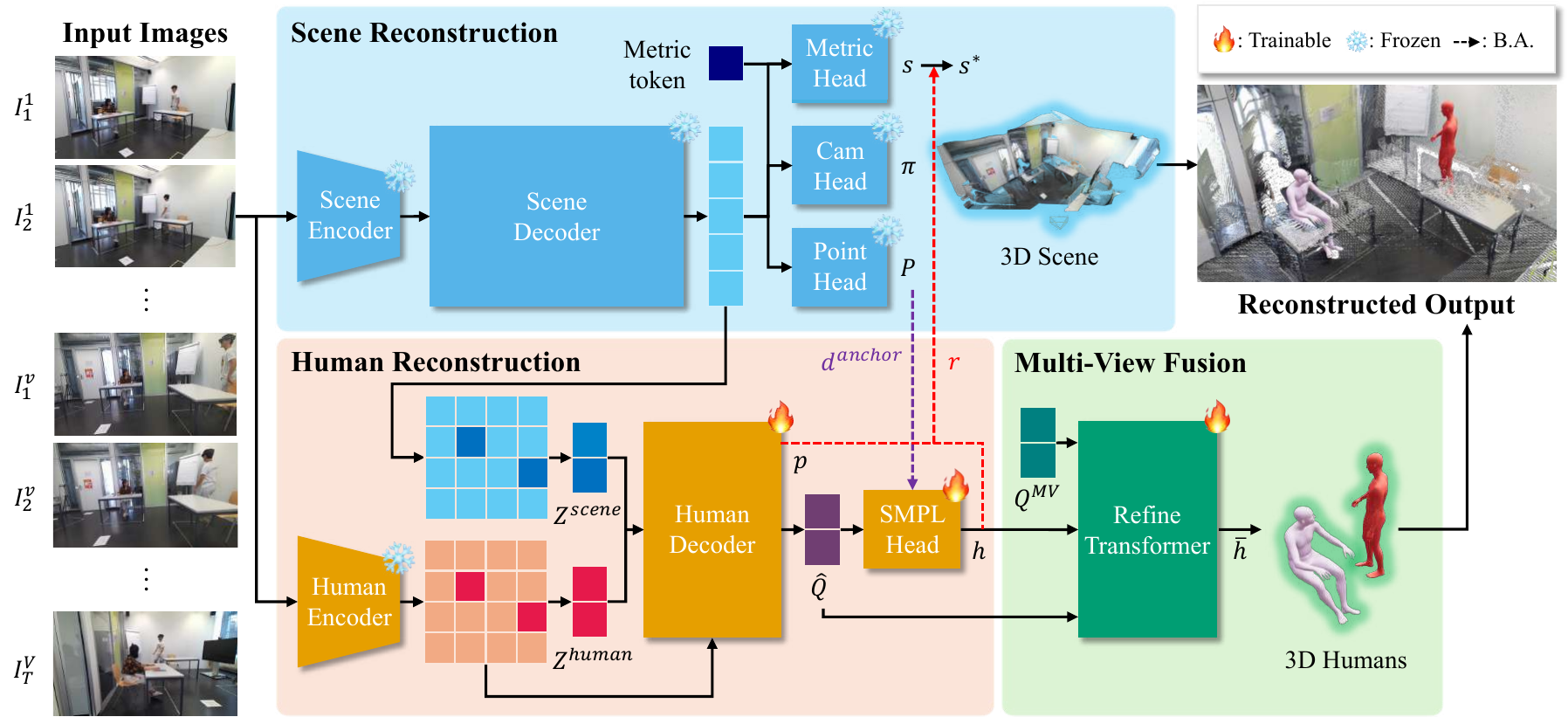}
  \caption{\textbf{Overview of our framework.} The scene encoder and decoder reconstruct cameras $\pi$, point maps $P$, and an approximate metric scale $s$. The human decoder and SMPL head predict per-view human estimates $h$. Bidirectional anchoring (B.A.) couples the two branches. Scene depth $d^{\text{anchor}}$ guides human placement, and the keypoint-based ratio $r$ calibrates the scene scale. For multi-view inputs, the multi-view fusion module merges per-view estimates into a global estimate $\bar{h}$.}
  \vspace{-4mm}
  \label{fig:pipeline}
\end{figure}

\section{Method}
\label{sec:method}
We present \textbf{\ours}, a feed-forward framework that reconstructs cameras, scene, and humans with identities from monocular or multi-view videos without test-time optimization. Given RGB frames $\{I_t^v\}_{t=1..T}^{v=1..V}$ from $V$ views and $T$ timesteps, the model estimates per-frame camera-space point maps $P_t^v$ and camera parameters $\pi_t^v$, as well as the SMPL-X parameters of $M$ people in a shared world coordinate system. For person $m$ at time $t$, the global human representation is
$\bar h_{t,m}=(\bar\theta_{t,m},\bar\beta_m,\bar R_{t,m},\bar\tau_{t,m})$,
where $\bar\theta_{t,m}\in\mathbb{R}^{52\times3}$ denotes canonical pose parameters, $\bar\beta_m\in\mathbb{R}^{10}$ shape parameters, $\bar R_{t,m}\in SO(3)$ global root rotation, and $\bar\tau_{t,m}\in\mathbb{R}^{3}$ global head translation. Bars denote final estimates in the world coordinate system. Fig.~\ref{fig:pipeline} shows the pipeline overview.

\subsection{Model Architecture}
\label{sec:model_architecture}
\ours~combines a scene prior and a human prior within a single architecture.
For human modeling, we adopt SMPL-X~\citep{SMPL-X:2019}, a parametric body model controlled by the pose parameters $\theta$ and the shape parameters $\beta$.

\vspace{2mm}\noindent\textbf{Scene Reconstruction.}
We use frozen Pi3X~\citep{wang2025pi} for scene reconstruction. We observe that human reconstruction can benefit from the geometric accuracy of the 3D foundation model, and fine-tuning the scene prior on human datasets can degrade scene quality.
We flatten the input frames into a sequence $\{I_n\}_{n=1..N}$, where $N=T\times V$ and process them with the scene encoder and decoder. The decoder jointly processes all frames with alternating attention and feeds its tokens to camera and point heads that predict $\pi_n$ and $P_n$. A metric token additionally predicts a global scale $s$, which is applied to the point maps and camera translations to obtain approximate metric-scale geometry. The decoded tokens are reshaped into feature maps $\{\hat F_n^{\mathrm{scene}}\}_{n=1}^N$ and passed to the human reconstruction branch.

\vspace{2mm}\noindent\textbf{Human Reconstruction.}
For human reconstruction, we adopt the frozen Multi-HMR~\citep{baradel2024multi} encoder for human-centric features, while the subsequent human decoder and SMPL head are trained from scratch. 
We note that the scene encoder alone does not capture fine-grained human details, and merging such heterogeneous representations into a single encoder requires dedicated distillation.
In parallel, the human encoder extracts human features $F_n^{\mathrm{human}}$ from each frame. Following~\citet{baradel2024multi}, a detection MLP identifies head patches, from which we extract a human token $Z_{n,m}^{\mathrm{human}}$ and the spatially corresponding scene token $Z_{n,m}^{\mathrm{scene}}$ for each detected person $m$. We concatenate these tokens with a positional embedding to form the person query $Q_{n,m}=[Z_{n,m}^{\mathrm{scene}}\mid Z_{n,m}^{\mathrm{human}}]$. The human decoder refines the queries through self-attention among people in the same frame and cross-attention to $F_n^{\mathrm{human}}$, producing $\hat Q_{n,m}$. Finally, the SMPL head predicts $(\theta_{n,m},\beta_{n,m},R_{n,m})$ from $\hat Q_{n,m}$, while the translation $\tau_{n,m}$ is derived from the scene depth map, as described in Sec.~\ref{sec:bidirectional_anchoring}.
\begin{figure}[!t]
  \centering
  \includegraphics[width=\linewidth]{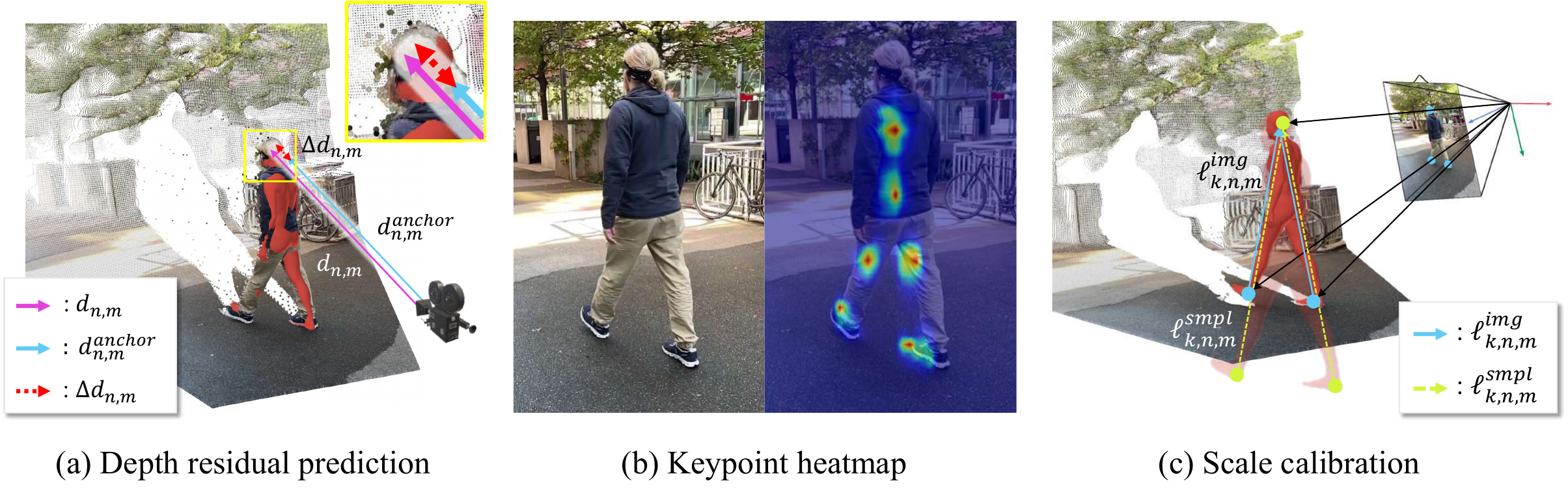}
  \caption{\textbf{Bidirectional Anchoring.} (a) Human depth is anchored on the head surface depth with a predicted residual to the head joint. (b) Keypoints for calibration are localized using heatmaps. (c) Scene scale is calibrated by comparing SMPL and image body proportions.}
  \label{fig:bidirectional_anchoring}
  \vspace{-4mm}
\end{figure}

\subsection{Bidirectional Anchoring}
\label{sec:bidirectional_anchoring}
As shown in Fig.~\ref{fig:teaser}b, aligning independently trained human and scene priors is challenging. If the predicted human depth is inconsistent, humans are placed too close to or too far from the camera, and if the scale is inconsistent, they float above or penetrate the ground. We observe that the two priors have complementary strengths. The scene prior provides consistent depth but only an approximate metric scale, whereas the human prior provides body geometry defined in metric units. Based on this observation, we introduce \emph{bidirectional anchoring}, which uses each prior to anchor the other.

\vspace{2mm}\noindent\textbf{Depth Residual Prediction.}
To place each human in the scene, we first estimate the 3D head joint translation $\tau_{n,m}\in\mathbb{R}^3$ in the camera coordinate system. We choose the head instead of the root because it is the most distinctive body part across individuals and is the least frequently occluded.
Instead of directly regressing the translation as in~\citet{chen2025human3r}, we predict the 2D head keypoint $p^{\text{head}}_{n,m}$ with an offset MLP from the detected head token $Z^{\text{human}}_{n,m}$, and unproject it with the predicted intrinsics $K_n\in\mathbb{R}^{3\times3}$ and the depth $d_{n,m}\in\mathbb{R}$ (Fig.~\ref{fig:bidirectional_anchoring}a).
We anchor this depth on the scene prior by sampling $d^{\text{anchor}}_{n,m}$ at the head keypoint from the depth map $D_n$, which is the z-coordinate of the point map $P_n$.
As the 3D head joint lies behind the visible head surface, we predict a positive residual $\Delta d_{n,m}$ from $\hat{Q}_{n,m}$ and add it to the anchor depth. The final 3D head translation is defined as
\begin{equation}
d_{n,m} = d^{\text{anchor}}_{n,m} + \Delta d_{n,m},
\hspace{5mm}
\tau_{n,m} = d_{n,m}\, K_n^{-1}
\begin{bmatrix} p^{\text{head}}_{n,m} \\ 1 \end{bmatrix}.
\label{eq:depth_residual}
\end{equation}
We observe that anchoring the depth on the scene prior stabilizes training and improves generalization compared to direct regression.

\vspace{2mm}\noindent\textbf{Scale Calibration.}
With Eq.~\ref{eq:depth_residual}, the SMPL head joints are precisely positioned in the scene. However, the scene prior estimates the scene at an \textit{approximate} metric scale, causing misalignment with metric-scale SMPL meshes.
We therefore calibrate the scene scale using the discrepancy between image-space 2D body proportions and those from the projected SMPL-X mesh.
We use the 2D head keypoint predicted by the offset MLP as the calibration reference, and select the spine, pelvis, knees, and feet as additional joints. For each non-head joint $k$, we first obtain an instance-conditioned heatmap by projecting the refined query $\hat{Q}_{n,m}$ with a joint-specific linear layer and taking its dot product with every patch token in the human feature map $F_n^{\mathrm{human}}$ (Fig.~\ref{fig:bidirectional_anchoring}b). 
Then, we calculate the 2D keypoint $p_{n,m}^{k}$ by applying soft-argmax to its heatmap. In addition, two linear heads predict a visibility and a confidence logit for each joint, indicating whether it is truncated or occluded, respectively.
We use an offset MLP instead of a heatmap for head localization, because it is more robust in crowded scenes as it predicts the head joint within the instance-specific head token boundary.
As shown in Fig.~\ref{fig:bidirectional_anchoring}c, we compute the image and SMPL head-to-joint lengths for each joint $k$ and derive the calibration ratio as
\begin{equation}
\ell^{\text{img}}_{k,n,m} = \| p^{\text{head}}_{n,m} - p^{k}_{n,m} \|_2,
\hspace{5mm}
\ell^{\text{smpl}}_{k,n,m} = \| \mathcal{J}^{\text{2D,head}}_{n,m} - \mathcal{J}^{\text{2D},k}_{n,m} \|_2
\end{equation}
\begin{equation}
r = \sum_{(k,n,m)\in\mathcal{S}} w^k_{n,m}\,
    \frac{\ell_{k,n,m}^\text{smpl}}{\ell_{k,n,m}^\text{img}}
\label{eq:calibration}
\end{equation}
where $\mathcal{J}^{\text{2D}}$ denotes the projected SMPL joint and $\mathcal{S}$ the set of joint-frame-person triplets used for calibration. The weight $w^k_{n,m}$ is the product of the predicted visibility and confidence, normalized to sum to one over $\mathcal{S}$, which makes unreliable joints contribute less to the ratio.
Finally, we obtain the calibrated scale as $s^* = r \cdot s$, enabling consistency between the scene and the SMPL meshes.
\subsection{Multi-View Fusion Module}
\label{sec:multi_view_fusion}
For multi-view human-scene reconstruction, existing approaches~\citep{muller2025reconstructing, rojas2025hamst3r, liu2026trophies} often rely on test-time optimization or precomputed cross-view correspondences, which are required for human and scene alignment. 
In contrast, our bidirectional anchoring already aligns humans and scene within each view, so per-view human estimates can be merged directly in a shared world coordinate system.
Building on this, we introduce a \emph{multi-view fusion module} that fuses per-view human estimates into a coherent global representation without test-time optimization or precomputed cross-view correspondences.

\vspace{2mm}\noindent\textbf{Component-wise Aggregation.}
The predicted human representation contains both view-invariant and view-dependent components. For each human, rather than uniformly averaging all per-view estimates, we aggregate each component according to its geometric properties. The canonical pose $\theta_{t,m}^v$ and body shape $\beta_{t,m}^v$ are shared across views, so we compute their weighted averages $(\theta^\text{avg}_{t,m},\beta^\text{avg}_{t,m})$ using reliability weights derived from the joint visibility and confidence estimates in Sec.~\ref{sec:bidirectional_anchoring}, which reduces sensitivity to occlusions. In contrast, the root rotation $R_{t,m}^v$ and 3D head translation $\tau_{t,m}^v$ are defined in each camera coordinate system, so we transform them into a shared world coordinate system. For the global root rotation $\bar R_{t,m}$, we compute a weighted average of the 6D representations and map the result back to $SO(3)$. For the global head translation $\bar \tau_{t,m}$, we observe that using multi-view ray triangulation instead of averaging translation enforces geometric consistency across views. Together, our component-wise aggregation strategies improve multi-view pose accuracy.

\begin{wrapfigure}{r}{0.26\textwidth}
  \vspace{-5mm}
  \centering
  \includegraphics[width=0.26\textwidth]{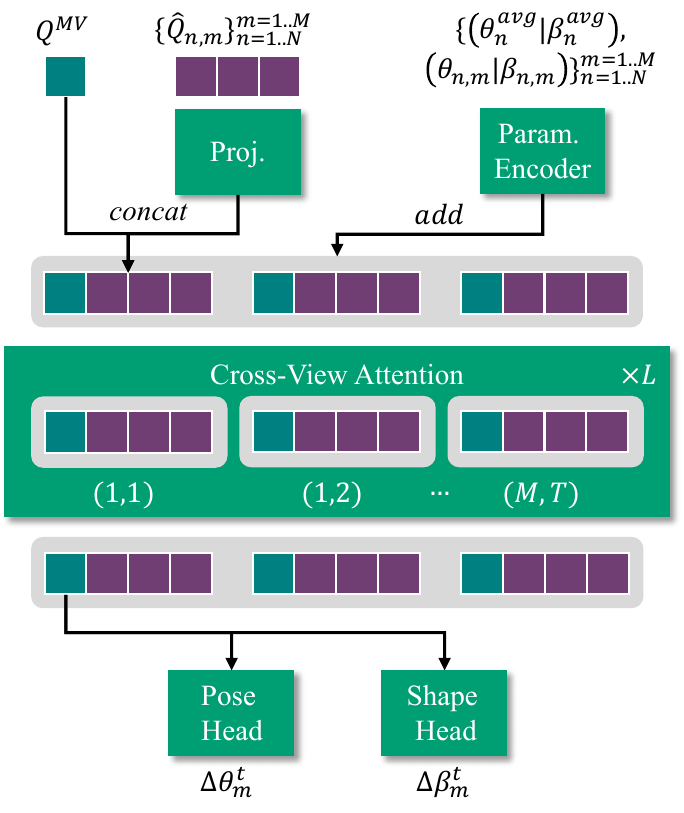}
  \vspace{-6mm}
  \caption{Refine TF.}
  \label{fig:refine_transformer}
\end{wrapfigure}

\vspace{2mm}\noindent\textbf{Refine Transformer.}
Component-wise aggregation provides robust initial estimates, but it does not fully exploit complementary information across views. We therefore introduce a \emph{refine transformer} that refines the averaged pose and shape estimates (Fig.~\ref{fig:refine_transformer}).
For each person $m$ at time $t$, we form a token sequence consisting of one learnable multi-view query $Q^{\mathrm{MV}}$ and the corresponding refined per-view queries $\hat Q_{t,m}^{v}$ from Sec.~\ref{sec:model_architecture}.
We add the embedding of the averaged and per-view parameters to the sequence and obtain the input sequence $[X^{\mathrm{MV}}_{t,m},X^{1}_{t,m},\ldots,X^{V}_{t,m}]$ as 
\begin{equation}
    X^\text{MV}_{t,m} = Q^\text{MV} + \operatorname{Enc}([\theta^\text{avg}_{t,m}|\beta^\text{avg}_{t,m}]),
\end{equation}
\begin{equation}
    X_{t,m}^v = \operatorname{Proj}(\hat Q_{t,m}^v) + \operatorname{Enc}([\theta_{t,m}^v|\beta_{t,m}^v]).
\end{equation}
The shared parameter encoder $\operatorname{Enc}(\cdot)$ maps concatenated pose and shape parameters to the token space, and $\operatorname{Proj}(\cdot)$ linearly projects each refined per-view query to the same dimension. 
The resulting sequence is processed by the refine transformer, where self-attention operates across the views of the same person at the same timestep. The multi-view output token of each sequence is decoded into pose and shape residuals, and the final pose and shape $(\bar\theta_{t,m}, \bar\beta_{m})$ are obtained as
\begin{equation}
    \bar{\theta}_{t,m} = \theta^{\text{avg}}_{t,m} + \Delta\theta_{t,m},
    \hspace{5mm}
    \bar{\beta}_{m} = \frac{1}{T}\sum_{t=1}^{T}\big(\beta^{\text{avg}}_{t,m} + \Delta\beta_{t,m}\big).
\end{equation}
\vspace{-4mm}

\vspace{2mm}\noindent\textbf{Geometric Multi-Person Association.}
Previous approaches rely on manually assigned identities~\citep{muller2025reconstructing}, which increase system complexity, or appearance-based ReID~\citep{li2024multi}, which is unreliable for visually similar individuals. We instead associate people using pose and global 3D position. 
Within each view, we form tracklets by matching refined query tokens $\hat{Q}_{t,m}^{v}$ across consecutive timesteps based on $\ell_2$ distance, and keep unmatched tokens in memory for re-tracking.
We construct a 2-nearest-neighbor view graph using mean camera centers. For each connected view pair, we compute matching costs between all tracklet pairs as
\begin{equation}
\mathcal{C}(a,b) = \frac{1}{|\mathcal{T}_{a,b}|}
\sum_{t \in \mathcal{T}_{a,b}} \frac{1}{J} \sum_{j=1}^{J}
\left(\lambda_p \left\|\mathcal{J}^{\text{3D},a}_{t,j}-\mathcal{J}^{\text{3D},b}_{t,j}\right\|_2
+
\lambda_{c}\left\|\mathcal{J}^{\text{3D},a,\text{} \text{canon}}_{t,j}-\mathcal{J}^{\text{3D},b,\text{canon}}_{t,j}\right\|_2\right).
\label{eq:matching_cost}
\end{equation}
where $\mathcal{T}_{a,b}$ denotes the set of overlapping timesteps, and $\mathcal{J}^{\text{3D},\mathrm{canon}}$ denotes joints without root rotation or translation. We set $\lambda_c=0.8$ and $\lambda_p=0.2$. For each view-graph edge, we use Hungarian matching to associate tracklets. We merge accepted matches across pairs using Union-Find, obtaining globally consistent identities. 
Unlike other components, our association is not learnable and runs only at inference, without updating any camera or human parameters.
Further details are provided in Appendix~\ref{sec:implementation}.
\section{Experiments}
\label{sec:experiments}
\subsection{Experimental Setup}
We train \ours~using BEDLAM~\citep{black2023bedlam}, 3DPW~\citep{von2018recovering}, MPII~\citep{andriluka20142d}, MSCOCO~\citep{lin2014microsoft}, and MammaSyn~\citep{cuevas2026mamma} datasets.
We evaluate our framework on two conventional tasks: (i) global human motion estimation and (ii) multi-view human pose estimation. 
We compare against methods with publicly available implementations or directly comparable results reported under the same established evaluation protocols, including Human3R~\citep{chen2025human3r} and HSfM~\citep{muller2025reconstructing}.
For global human motion estimation, we evaluate on EMDB-2~\citep{kaufmann2023emdb}, which contains monocular videos captured by dynamic cameras, and RICH~\citep{huang2022capturing}, which provides fixed multi-view recordings. Following the evaluation protocol of Human3R~\citep{chen2025human3r}, we report WA-MPJPE and W-MPJPE, which measure mean joint errors after aligning each 100-frame segment using the entire segment and only its first two frames, respectively, and RTE, which measures the normalized root translation error after SE(3) alignment.
For multi-view human pose estimation, we evaluate on EgoHumans~\citep{khirodkar2023ego} and EgoExo4D~\citep{grauman2024ego} under the single-timestep evaluation protocol of HSfM~\citep{muller2025reconstructing}. We report W-MPJPE$^\dagger$ using camera-only SE(3) alignment, GA-MPJPE after jointly aligning all people, and PA-MPJPE after person-wise Procrustes alignment. 
Implementation details are provided in Appendix~\ref{sec:implementation}.

    \subsection{Human and Scene Reconstruction}
\begin{table}[t]
\centering
\caption{Quantitative comparison on RICH and EMDB-2 datasets. Ours-multi uses all available views and is reported for reference. \colorbox{red!20}{red}: the best, \colorbox{yellow!50}{yellow}: the second best.}
\resizebox{\textwidth}{!}{
\begin{tabular}{l|ccc|ccc|ccc}
\toprule
& \multicolumn{3}{c|}{\textbf{Settings}} & \multicolumn{3}{c|}{\textbf{RICH}} & \multicolumn{3}{c}{\textbf{EMDB-2}} \\
& Multi-View & Multi-Person & Scene
& WA-MPJPE$\downarrow$ & W-MPJPE$\downarrow$ & RTE(\%)$\downarrow$
& WA-MPJPE$\downarrow$ & W-MPJPE$\downarrow$ & RTE(\%)$\downarrow$ \\
\midrule
JOSH3R     & \xmark & \checkmark & \checkmark & -     & -     & -   & 220.0 & 661.7 & 13.1 \\
UniSH      & \xmark & \xmark & \checkmark & 118.1 & \cellcolor{yellow!50}183.2 & 4.8 & 118.5 & 270.1 & 5.8 \\
Human3R    & \xmark & \checkmark & \checkmark & \cellcolor{yellow!50}110.0 & 184.9 & \cellcolor{yellow!50}3.3 & \cellcolor{yellow!50}112.2 & \cellcolor{yellow!50}267.9 & \cellcolor{yellow!50}2.2 \\
\midrule
Ours-mono  & \xmark & \checkmark & \checkmark & \cellcolor{red!20}84.9    & \cellcolor{red!20}132.4    & \cellcolor{red!20}2.6
           & \cellcolor{red!20}83.6    & \cellcolor{red!20}214.4    & \cellcolor{red!20}1.4 \\
\midrule
Ours-multi & \checkmark & \checkmark & \checkmark & \textbf{45.4} & \textbf{61.2} & \textbf{0.9}   & -    & -    & -   \\
\bottomrule
\end{tabular}
}
\label{tab:global_motion}
\end{table}
\vspace{2mm}\noindent\textbf{Global Motion Estimation.} We compare our framework with feed-forward monocular human-scene reconstruction methods on EMDB-2 and RICH. 
Among the baselines, JOSH3R~\citep{liu2025joint} requires 2D human masks, UniSH~\citep{li2026unish} relies on bounding-box detection and supports only a single person, and Human3R~\citep{chen2025human3r} supports multiple people but only monocular inputs.
As shown in Table~\ref{tab:global_motion}, our framework achieves lower global motion errors than these baselines in the monocular setting, while using multiple views further improves accuracy.

\begin{table}[t]
\centering
\footnotesize
\caption{Quantitative comparison on EgoExo4D and EgoHumans datasets. For fair comparison, all methods are evaluated with ground-truth identities. All metrics are in meters.}
\resizebox{\textwidth}{!}{
\begin{tabular}{l|ccc|cc|ccc|c}
\toprule
& \multicolumn{3}{c|}{\textbf{Settings}} & \multicolumn{2}{c|}{\textbf{EgoExo4D}} & \multicolumn{4}{c}{\textbf{EgoHumans}} \\
& Video & No ReID & No Optim.
& W-MPJPE$^{\dagger}\downarrow$ & PA-MPJPE$\downarrow$
& W-MPJPE$^{\dagger}\downarrow$ & GA-MPJPE$\downarrow$ & PA-MPJPE$\downarrow$ & Runtime$\downarrow$ \\
\midrule
HSfM (init.) & \xmark & \xmark & \checkmark & 5.29 & \cellcolor{yellow!50}0.07 & 4.28 & 0.51 & \cellcolor{yellow!50}0.06 & \cellcolor{yellow!50}$\sim$21s \\
HSfM    & \xmark & \xmark & \xmark & 0.56 & \cellcolor{red!20}0.06 & \cellcolor{yellow!50}1.04 & \cellcolor{yellow!50}0.21 & \cellcolor{red!20}0.05 & $\sim$118s \\
HAMSt3R & \xmark & \checkmark & $\triangle$ & \cellcolor{yellow!50}0.51 & 0.09 & 3.80 & 0.42 & 0.14 & $\sim$32s \\
\midrule
Ours    & \checkmark & \checkmark & \checkmark & \cellcolor{red!20}0.23 & \cellcolor{red!20}0.06
        & \cellcolor{red!20}0.37 & \cellcolor{red!20}0.20 & \cellcolor{red!20}0.05 & \cellcolor{red!20}$\sim$1.1s \\
\bottomrule
\end{tabular}
}
\label{tab:mv_pose}
\end{table}

\begin{table}[t]
\centering
\scriptsize
\setlength{\tabcolsep}{4pt}
\begin{minipage}[t]{0.5\linewidth}
  \centering
  \caption{Ablation on multi-view fusion strategies.}
  \label{tab:mv_fusion}
  \begin{tabular}{l|cccc}
  \toprule
  RICH-MV & WA-M.$\downarrow$ & W-M.$\downarrow$ & RTE(\%)$\downarrow$ & PA-M.$\downarrow$ \\
  \midrule
  Ours-mono   & 84.9 & 132.4 & 2.6 & 51.0 \\
  \midrule
  (1) Avg.   & 65.0  & 93.3  & 1.2 & 38.2\\
  (2) Tri.+Avg.   & 48.2 & 67.0 & 1.0 & 38.2\\
  (3) Tri.+Conf.   & 47.8  & 66.2  & 1.0 & 37.4\\
  \rowcolor{gray!15}
  (4) Tri.+Conf.+Refine & \textbf{45.4} & \textbf{61.2}& \textbf{0.9} & \textbf{35.6}\\
  \bottomrule
  \end{tabular}
\end{minipage}\hfill
\begin{minipage}[t]{0.45\linewidth}
  \centering
  \vspace{0pt}
  \includegraphics[width=\linewidth]{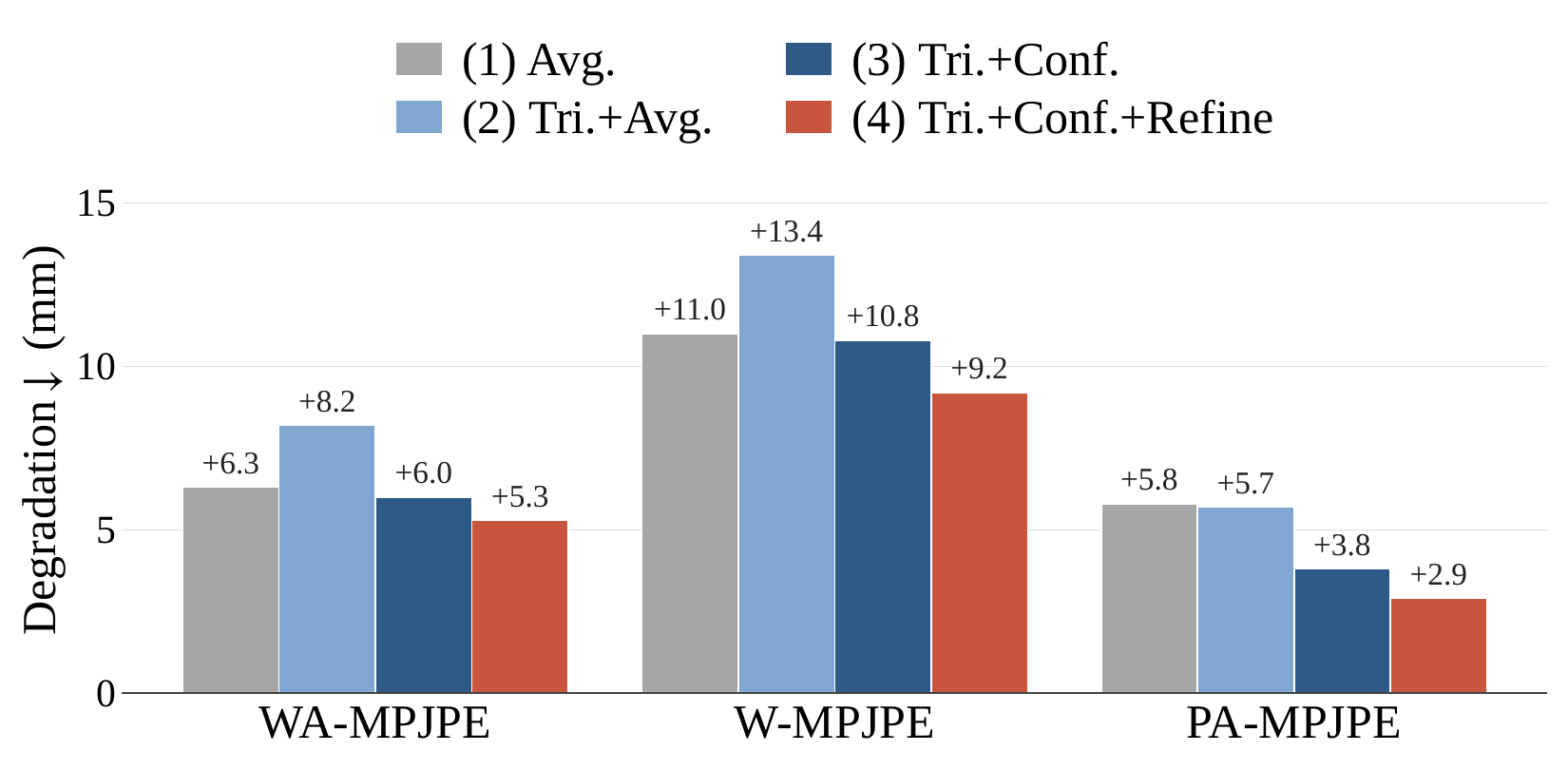}
  \captionof{figure}{Sensitivity analysis.}
  \label{fig:sensitivity}
  \vspace{-5mm}
\end{minipage}
\end{table}

\vspace{2mm}\noindent\textbf{Multi-View Pose Estimation.} We compare our framework with existing multi-view human-scene reconstruction methods under the single-timestep protocol used by HSfM~\citep{muller2025reconstructing} and HAMSt3R~\citep{rojas2025hamst3r}.
We do not include Trophies~\citep{liu2026trophies}, as its multi-view evaluation follows a different protocol and its implementation is not publicly available.
Unlike ours, Trophies relies on test-time optimization with external metric depth, 2D keypoints, and human masks, and does not address cross-view identity association.
As shown in Table~\ref{tab:mv_pose}, our method reduces global pose error while achieving competitive local pose accuracy without test-time optimization. On a single EgoHumans scene (three people and four views) using one NVIDIA V100 GPU, it runs over $100\times$ faster than HSfM and $28\times$ faster than HAMSt3R.
\begin{wraptable}{r}{0.4\linewidth}
\centering
\scriptsize
\vspace{-2mm}
\setlength{\tabcolsep}{4pt}
\caption{Ablation on scale calibration. Both metrics are computed in meters.}
\label{tab:scale_calib}
\begin{tabular}{l|cc|cc}
\toprule
& \multicolumn{2}{c|}{\textbf{EgoExo4D}}
& \multicolumn{2}{c}{\textbf{EgoHumans}} \\
& W-M.$^{\dagger}\downarrow$ & TE.$\downarrow$
& W-M.$^{\dagger}\downarrow$ & TE.$\downarrow$ \\
\midrule
Ours-$s$   
& 0.31 & 0.54 & 1.56 & 3.01 \\
\rowcolor{gray!15}
Ours-$s^*$ 
& \textbf{0.23} & \textbf{0.19} & \textbf{0.37} & \textbf{0.31} \\
\bottomrule
\end{tabular}
\vspace{-\baselineskip}
\end{wraptable}
Finally, we compare applying the calibration ratio $r$ to the scene scale (Ours-$s^*$) against keeping the scene scale fixed and instead rescaling the SMPL meshes (Ours-$s$). 
We report W-MPJPE$^\dagger$ and mean camera translation error (TE), both computed after only a rigid SE(3) alignment.
As both metrics are sensitive to metric scale, Table~\ref{tab:scale_calib} shows that calibrating the scene using the human prior improves the estimated scene scale.

\begin{wraptable}{r}{0.45\linewidth}
\vspace{-\baselineskip}
\centering
\scriptsize
\setlength{\tabcolsep}{4pt}
\caption{Camera pose and depth accuracy.}
\label{tab:scene_eval}
\begin{tabular}{l|cc|cc}
\toprule
\multirow{2}{*}{}
  & \multicolumn{2}{c|}{\textbf{Camera (EMDB-2)}}
  & \multicolumn{2}{c}{\textbf{Depth (Bonn)}} \\
  & ATE$\downarrow$ & RPE$_\text{t}\downarrow$
  & AbsRel$\downarrow$ & $\delta{<}1.25\uparrow$ \\
\midrule
CUT3R                       & .059          & .013          & .048          & .970 \\
Human3R  & .037          & .012          & .050          & .961 \\
\midrule
Pi3                         & \textbf{.025} & .010          & \textbf{.026}          & .985 \\
UniSH    & .121          & .021          & .034          & .985 \\
\midrule
Pi3X                        & .026          & \textbf{.008} & \textbf{.026}          & \textbf{.986} \\
\rowcolor{gray!15} Ours & .026      & \textbf{.008} & \textbf{.026} & \textbf{.986} \\
\bottomrule
\end{tabular}
\vspace{-\baselineskip}
\end{wraptable}

\vspace{2mm}\noindent\textbf{Scene Reconstruction.} We measure camera accuracy on EMDB-2 with the absolute trajectory error (ATE) and the relative pose error (RPE$_t$), and depth accuracy on Bonn~\citep{palazzolo2019iros} with the absolute relative error (AbsRel) and the thresholded accuracy ($\delta<1.25$). As shown in Table~\ref{tab:scene_eval}, our framework preserves the accuracy of the scene prior, whereas UniSH~\citep{li2026unish}, which fine-tunes its backbone, degrades camera accuracy. 
This supports our frozen scene design choice in Sec.~\ref{sec:model_architecture}.

\subsection{Analysis}

\vspace{2mm}\noindent\textbf{Fusion Components.} 
We ablate multi-view fusion components on RICH using PA-MPJPE to evaluate local human quality. As shown in Table~\ref{tab:mv_fusion}, uniform averaging (Avg.) performs worst, whereas replacing translation averaging with ray triangulation (Tri.) yields the largest gain. Confidence weighting (Conf.) further improves accuracy, and the refine transformer (Refine) enhances local pose. 
Fig.~\ref{fig:sensitivity} evaluates robustness to unreliable views by occluding a person of a randomly selected view by 60\% with a gray box, as in \citet{kocabas2021pare}. Triangulation alone is most sensitive to such views, while confidence weighting suppresses unreliable views and reduces the degradation. The refine transformer further lowers it, validating our fusion design choice.

\begin{wraptable}{r}{0.42\linewidth}
\vspace{-5mm}
\centering
\scriptsize
\setlength{\tabcolsep}{4pt}
\caption{Ablation on alignment strategies. All variants are trained up to Stage~1.}
\label{tab:align_ablation}
\begin{tabular}{l|ccc}
\toprule
EMDB-2 & WA-M.$\downarrow$ & W-M.$\downarrow$ & RTE(\%)$\downarrow$ \\
\midrule
w/o scale calib. & 157.8 & 422.3 & 4.2 \\
\midrule
Direct transl. & 100.9 & 251.4 & 1.9 \\
Direct depth & 111.8 & 276.4 & 2.2 \\
\midrule
\rowcolor{gray!15}
Ours-Stage~1 & \textbf{84.7} & \textbf{216.4} & \textbf{1.4} \\
\bottomrule
\end{tabular}
\vspace{-15mm}
\end{wraptable}
\vspace{2mm}\noindent\textbf{Bidirectional Anchoring.} We evaluate the effect of scale and depth alignment in bidirectional anchoring on EMDB-2 (Table~\ref{tab:align_ablation}). Without scale calibration, errors increase substantially due to the mismatch between two priors. For depth alignment, direct depth performs worst, while direct translation shows better accuracy. Our depth residual prediction performs best.

\begin{wraptable}{r}{0.45\linewidth}
\centering
\scriptsize
\setlength{\tabcolsep}{4pt}
\caption{Ablation on association weights.}
\label{tab:mp_assoc}
\begin{tabular}{cc|cccc}
\toprule
$\lambda_c$ & $\lambda_p$ & Acc.(\%) & F1(\%) & Prec.(\%) & Recall(\%) \\
\midrule
0.0 & 1.0 & 89.5 & 78.9 & \textbf{94.1} & 67.9 \\
0.2 & 0.8 & 89.9 & 79.9 &  94.0 & 69.6 \\
0.5 & 0.5 & 90.7 & 82.1 &  92.8 & 73.6 \\
\rowcolor{gray!15}
0.8 & 0.2 & \textbf{90.9} & \textbf{83.3} &  88.5 & 78.7 \\
1.0 & 0.0 & 68.3 & 59.6 & 47.2 & \textbf{80.8} \\
\bottomrule
\end{tabular}
\vspace{-\baselineskip}
\end{wraptable}

\vspace{2mm}\noindent\textbf{Association Ablation.}
We evaluate multi-person association on EgoHumans using ReID accuracy, precision, and recall in Table~\ref{tab:mp_assoc}. As reported in~\citet{muller2025reconstructing}, appearance-based ReID achieves $51.2\%$ accuracy, showing the difficulty of distinguishing similar individuals. Our geometric association substantially improves accuracy to $90.9\%$. Pose alone yields the highest recall but the lowest precision by over-matching similar poses. Combining pose and position performs best, with $\lambda_c=0.8$ and $\lambda_p=0.2$, while accuracy remains stable for $\lambda_c \in [0.2, 0.8]$.

\begin{figure}[t]
  \centering
  \includegraphics[width=1.0\linewidth]{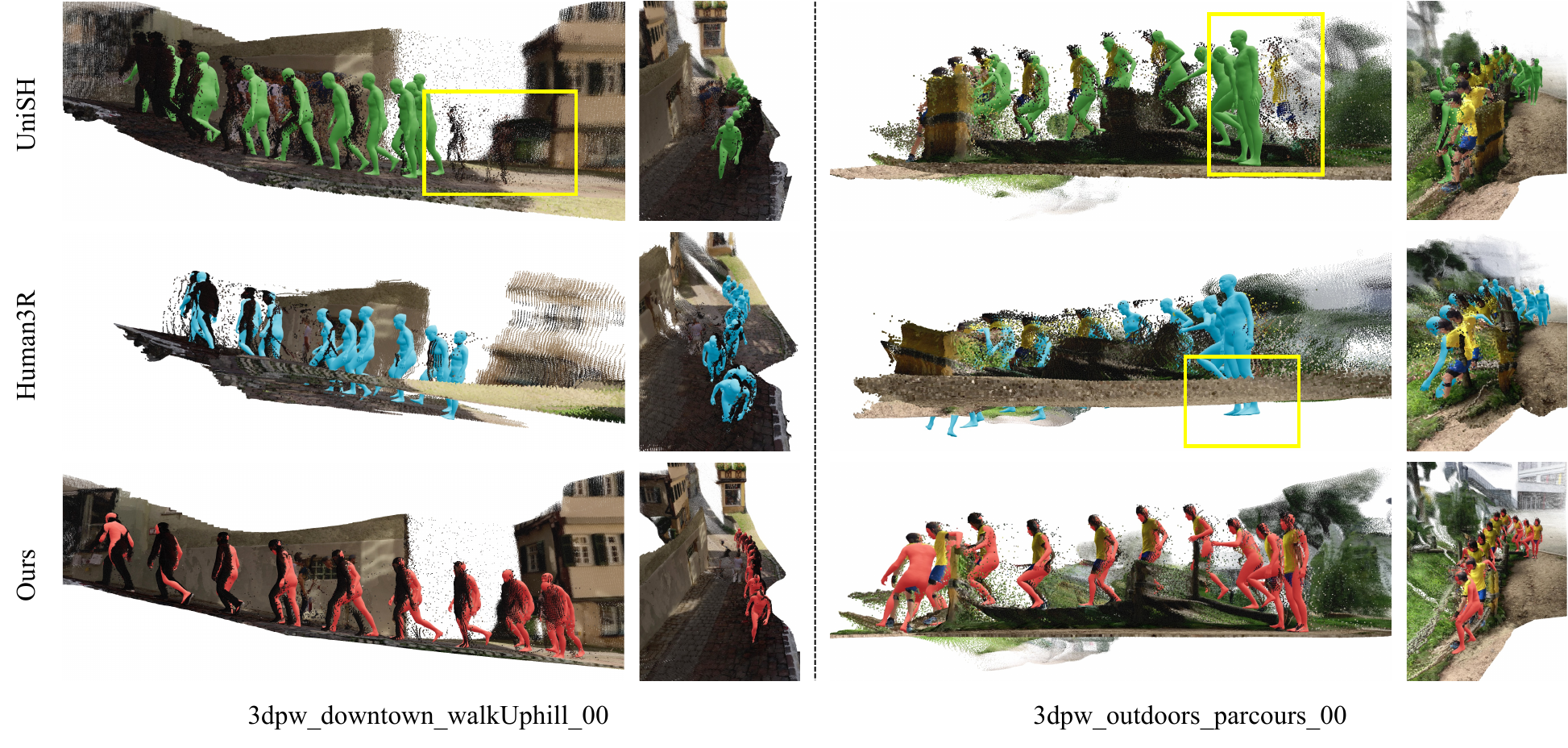}
  \caption{\textbf{Qualitative results on the 3DPW dataset.} We compare our method with monocular human-scene reconstruction baselines, including UniSH and Human3R.}
  \label{fig:qual_comparison}
\end{figure}

\vspace{2mm}\noindent\textbf{Qualitative Analysis.} 
We compare qualitative results with other monocular human–scene reconstruction methods on the 3DPW dataset. As shown in Fig.~\ref{fig:qual_comparison}, UniSH shows depth misalignment between the scene point cloud and human mesh. In addition, Human3R produces penetrations when the estimated scene and human scales are inconsistent. In contrast to these baselines, our method achieves coherent human–scene alignment. Additional results are provided in Appendices~\ref{sec:additional_analysis} and \ref{sec:results}.
\section{Conclusion}
In this paper, we present \ours, a feed-forward framework that reconstructs cameras, scene, and humans with identities from monocular or multi-view video without test-time optimization. We introduce \emph{bidirectional anchoring} to bridge complementary human and scene priors: human keypoints calibrate scene scale to match the human body, while scene depth guides global human placement. Building on the per-view alignment, our \emph{multi-view fusion module} combines confidence-weighted averaging, triangulation, a refine transformer, and geometry-based association to handle unreliable views and visually similar individuals. Experiments demonstrate improved global motion and multi-view pose accuracy by up to 28\% and 64\% over previous human-scene reconstruction methods, while running over 28$\times$ faster than optimization-based multi-view approaches.

\vspace{2mm}\noindent\textbf{Limitations and Future Work.}
\ours~shares a common limitation with the CenterNet-based paradigm in that it cannot recover when the head is fully invisible in the monocular setting.
In addition, since scale calibration uses the human body, it may be biased for people whose body sizes deviate from the training distribution, such as children.
We plan to extend the work by distilling heterogeneous priors into a unified encoder.

\subsection*{AI use statement}

In this work, we used generative AI tools to assist with implementing the method (writing and debugging parts of the model, training, and evaluation code), to provide feedback on the design of experiments, and to help interpret experimental results.
We have not used generative AI tools to generate
synthetic datasets, develop theoretical models or conceptual frameworks, formulate mathematical claims, provide critical ingredients for or assist in writing proofs, propose or refine hypotheses, or clean and reformat datasets; translation and qualitative data analysis are not applicable to this work.
Additionally, we used generative AI tools to draft parts of the paper, to edit the manuscript for grammar and readability, to suggest the paper's structure, title, and keywords, to search for and summarize related literature and to verify and format references, to suggest experimental parameters, and to write auxiliary code such as visualization and data-processing scripts.
We have reviewed all AI-assisted work.
All AI-assisted code was reviewed and tested for correctness by the authors, and all reported results were produced by running the authors' verified code.
AI feedback on experimental design and AI-suggested interpretations of results were critically assessed by the authors, who made all final decisions on the experiments and the conclusions drawn from them.
AI-drafted text was rewritten and checked by the authors, and all AI-identified literature was verified against the original sources.
We take responsibility for the final content of this work, including text, claims or artifacts produced with the aid of generative AI.

\subsection*{Ethics statement}

This work does not involve the collection of new human-subject data; all experiments use publicly available datasets under their respective licenses.
Since our method reconstructs and tracks multiple people across cameras, it could potentially be misused for surveillance without consent.
We encourage its use only with the informed consent of captured individuals and in compliance with applicable privacy regulations.

\subsection*{Reproducibility statement}

The method is described in Sec.~\ref{sec:method}, and full details on architecture, training schedule, losses, hyperparameters, and evaluation protocols are provided in Appendix~\ref{sec:implementation}.
All datasets used are publicly available.
We will release our code and pretrained models upon publication.



\bibliography{iclr2027_conference}
\bibliographystyle{iclr2027_conference}

\clearpage
\appendix
\begin{center}
    \LARGE \textbf{Appendix}
\end{center}
\vspace{5mm}
In this section, we provide the implementation details of our framework, \ours, along with the additional analyses and results to support our claims. In Sec.~\ref{sec:additional_analysis}, we present additional analyses of our framework. In Sec.~\ref{sec:implementation}, we describe the implementation details. In Sec.~\ref{sec:results}, we present additional results across diverse scenarios, as well as failure cases.

\section{Additional Analysis}
\label{sec:additional_analysis}
\subsection{Bidirectional Anchoring}
To further analyze the effect of our bidirectional anchoring, we perform physical plausibility experiment, trajectory comparison and visualize heatmaps on truncated and occluded scenarios.

\begin{wraptable}{r}{0.3\textwidth}
\vspace{-\baselineskip}
\centering
\scriptsize
\caption{Physical plausibility evaluation.}
\resizebox{\linewidth}{!}{
\begin{tabular}{l | cc}
\toprule
RICH & Penet.$\downarrow$ & Float$\downarrow$ \\
\midrule
UniSH  & 13.7 & 11.5 \\
Human3R & 5.3 & 18.1 \\
\midrule
\rowcolor{gray!15}
Ours    & \textbf{5.0} & \textbf{7.6} \\
\bottomrule
\end{tabular}}
\label{tab:physical}
\vspace{-\baselineskip}
\end{wraptable}
\vspace{2mm}\noindent\textbf{Physical Plausibility.}
We measure physical plausibility on RICH~\citep{huang2022capturing}, where calibrated cameras and scene scans are provided. We define contact vertices as ground-truth body vertices within 3\,cm of the scan. Each predicted body is transformed into the scan frame using the ground-truth camera pose, and the corresponding vertices are compared with the local contact surface. Following HUMOS~\citep{tripathi2024humos}, we report penetration and float as their mean distances (cm) below and above the surface. As shown in Table~\ref{tab:physical}, our bidirectional anchoring achieves lower penetration and float than UniSH~\citep{li2026unish} and Human3R~\citep{chen2025human3r}, which demonstrates its effectiveness in placing humans coherently within the scene.

\begin{figure}[h]
  \centering
  \includegraphics[width=\linewidth]{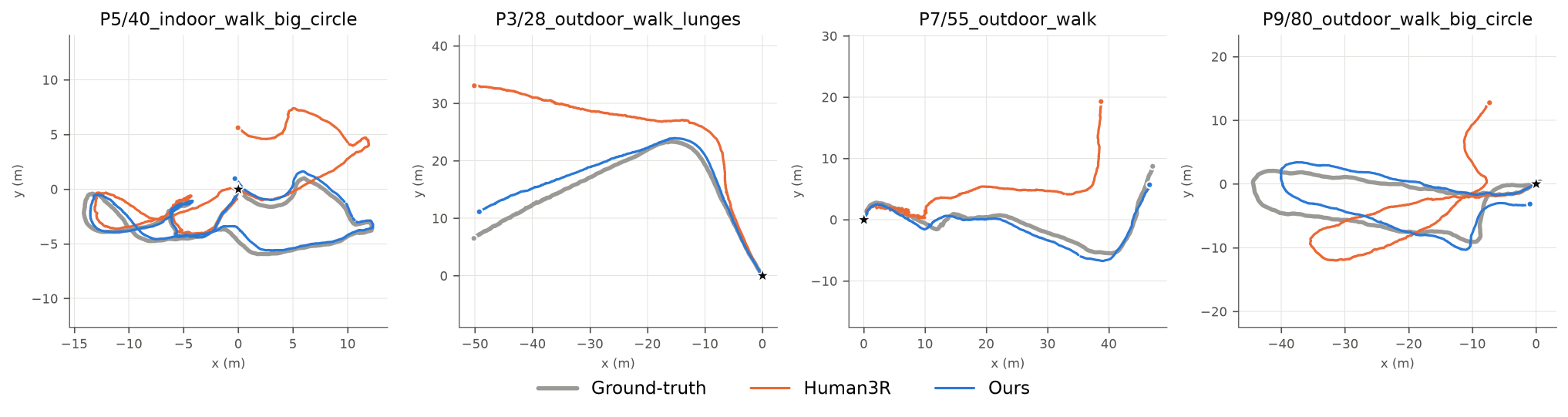}
  \caption{Trajectory comparison.}
  \label{fig:trajectory}
\end{figure}

\vspace{2mm}\noindent\textbf{Trajectory Comparison.}
In Fig.~\ref{fig:trajectory}, we compare our global root trajectories with Human3R, the previous state-of-the-art method. We select four EMDB-2 sequences with long walking paths, and visualize from a top-down view with the ground truth. Each trajectory is aligned to the ground truth once, using its first 100 frames. Human3R gradually drifts away from the ground truth as the sequence progresses, whereas our trajectories closely follow it in both shape and scale over the entire sequence. This shows that bidirectional anchoring yields globally consistent human motion.

\begin{figure}[t]
  \centering
  \includegraphics[width=\linewidth]{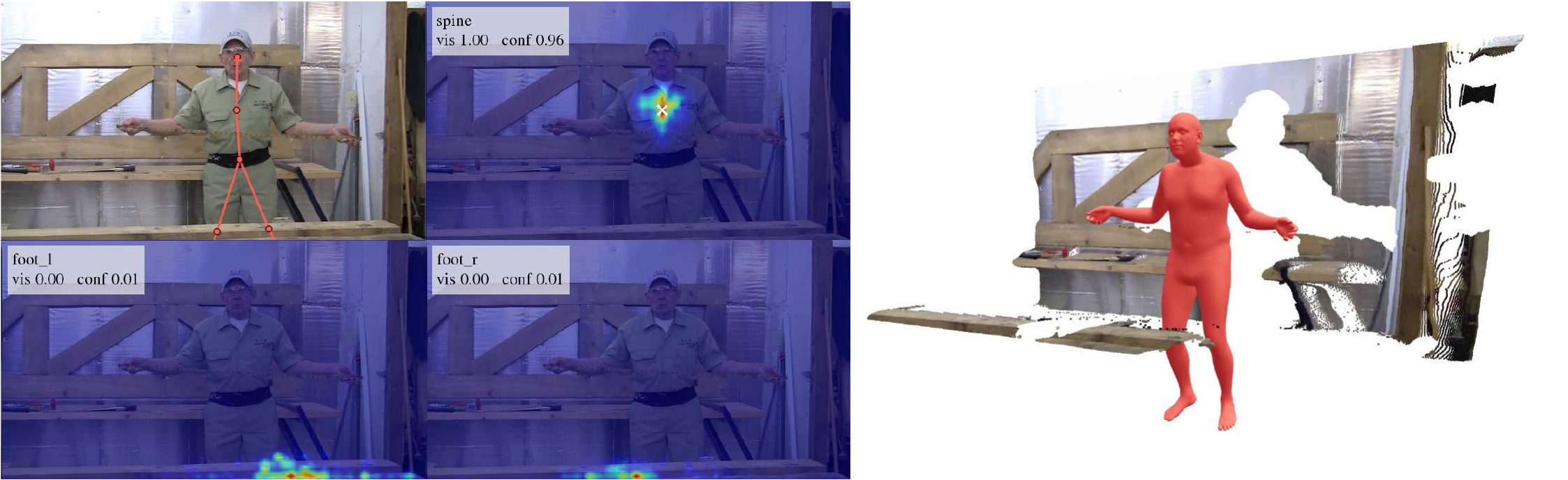}
  \caption{Heatmap visualization of truncated scene and the reconstructed output.}
  \label{fig:heatmap_truncated}
\end{figure}

\begin{figure}[t]
  \centering
  \includegraphics[width=\linewidth]{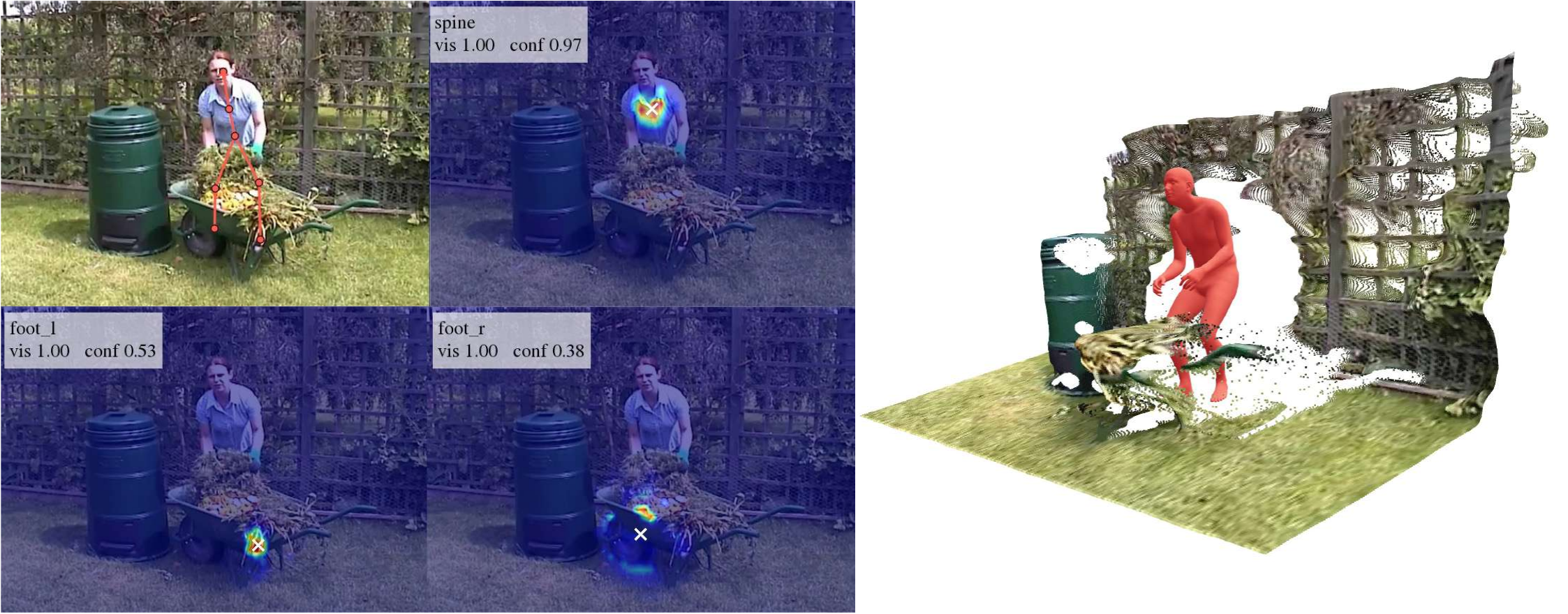}
  \caption{Heatmap visualization of occluded scene and the reconstructed output.}
  \label{fig:heatmap_occluded}
\end{figure}

\vspace{2mm}\noindent\textbf{Heatmap Visualization.}
We analyze how our bidirectional anchoring behaves under truncation and occlusion by visualizing the 2D keypoints and the heatmaps of the spine and both feet. When the lower body is truncated by the image boundary (Fig.~\ref{fig:heatmap_truncated}), the predicted visibility of both feet converges to zero. When the feet are inside the image but occluded (Fig.~\ref{fig:heatmap_occluded}), the visibility remains close to 1.0 while the confidence decreases. Since the calibration weight $w^k_{n,m}$ in Eq.~\ref{eq:calibration} is defined as the product of visibility and confidence, truncated and occluded joints contribute less to the calibration ratio, which allows the scene scale to be calibrated robustly in both cases. As shown in both Fig.~\ref{fig:heatmap_truncated} and Fig.~\ref{fig:heatmap_occluded}, both humans are placed coherently in the scene.

\subsection{Robustness Analysis}
We present qualitative results of our framework across challenging scenarios, including in-the-wild scenes with truncated, occluded, or crowded subjects, multi-view settings, and monocular sequences with temporal occlusions.

\begin{figure}[t]
  \centering
  \includegraphics[width=\linewidth]{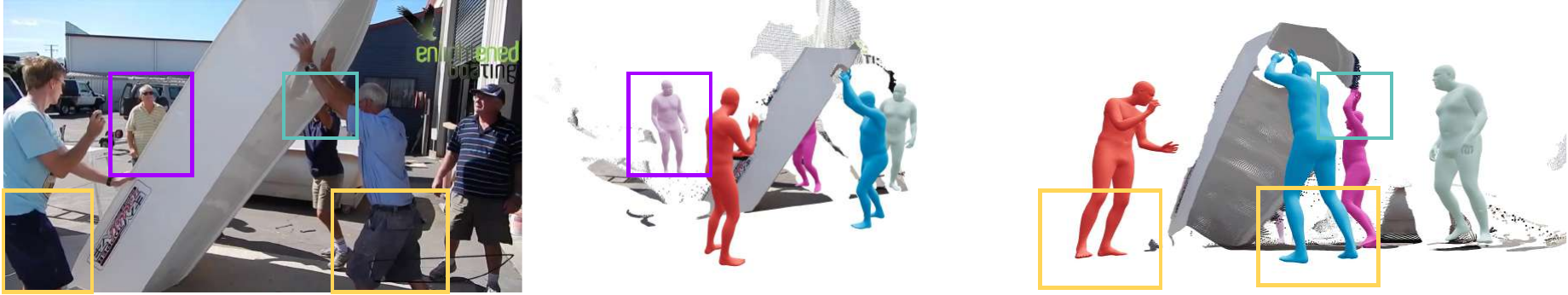}
  \caption{Challenging scenario with truncation, occlusion, and object in front of the head.}
  \label{fig:challenging}
\end{figure}

\begin{figure}[t]
  \centering
  \includegraphics[width=\linewidth]{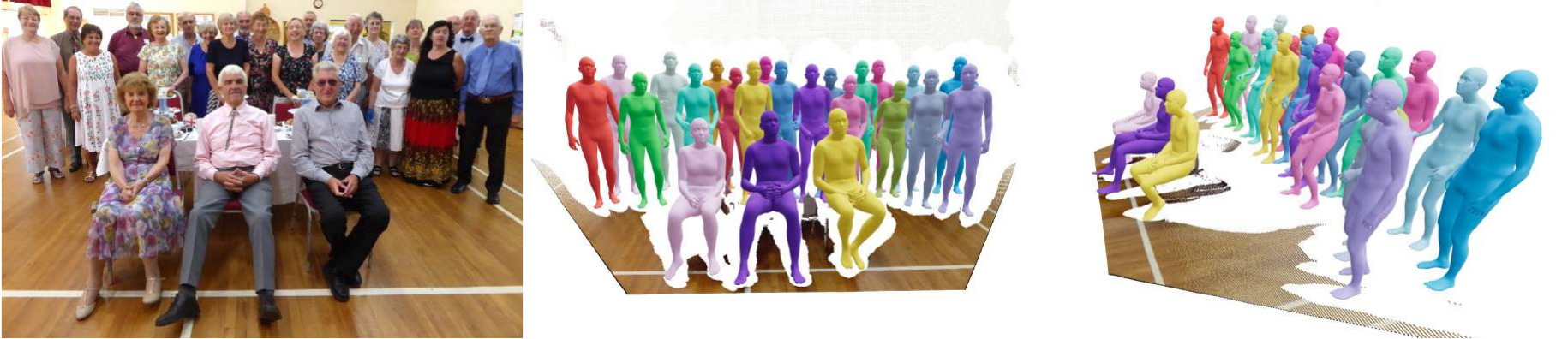}
  \caption{Challenging scenario with crowded scene.}
  \label{fig:crowded}
\end{figure}

\vspace{2mm}\noindent\textbf{In-the-wild Scenarios.}
We show qualitative results in challenging in-the-wild scenarios in Fig.~\ref{fig:challenging} and Fig.~\ref{fig:crowded}. Fig.~\ref{fig:challenging} includes individuals with truncation (yellow box), occlusion (purple box), and objects placed in front of the head (cyan box). Our framework is able to recover the truncated and occluded body parts, while robustly aligning the humans with the scene. Also, our framework is able to infer correct depth even when the head is partially covered by an object. In Fig.~\ref{fig:crowded}, we increase the maximum number of people $M$ from 10 to 30 at inference, and reconstruct the crowded scene. Although trained with at most 10 people per frame, our framework successfully recovers more than 20 people. Moreover, our depth residual prediction preserves the relative depth ordering among people.

\begin{figure}[t]
  \centering
  \includegraphics[width=0.9\linewidth]{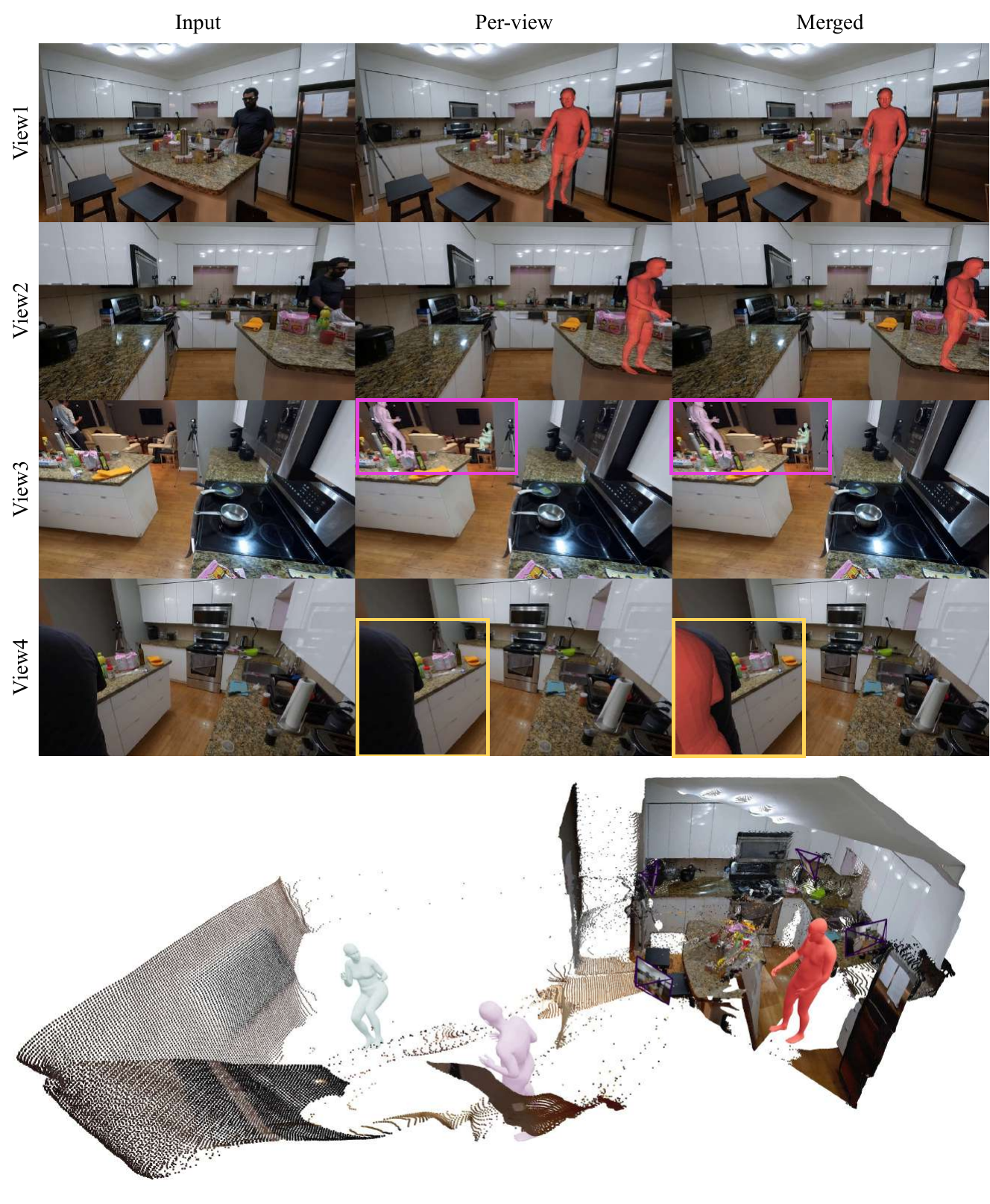}
  \caption{Multi-view reconstruction process.}
  \label{fig:mv_process}
\end{figure}

\vspace{2mm}\noindent\textbf{Multi-View Scenario.}
We illustrate how per-view estimates are merged into the global representation in multi-view scenario of EgoExo4D~\citep{grauman2024ego}. In Fig.~\ref{fig:mv_process}, we show the multi-view reconstruction process of a scene with 3 people and 4 views. Our framework is able to reconstruct global humans even when detection fails in View~4 (yellow box). Furthermore, two people are observed only in View~3 (pink box) and are still successfully placed in the world coordinate system even though they are not visible in any other view. This demonstrates the practical applicability of our framework, which requires only a single visible view.

\begin{figure}[t]
  \centering
  \includegraphics[width=\linewidth]{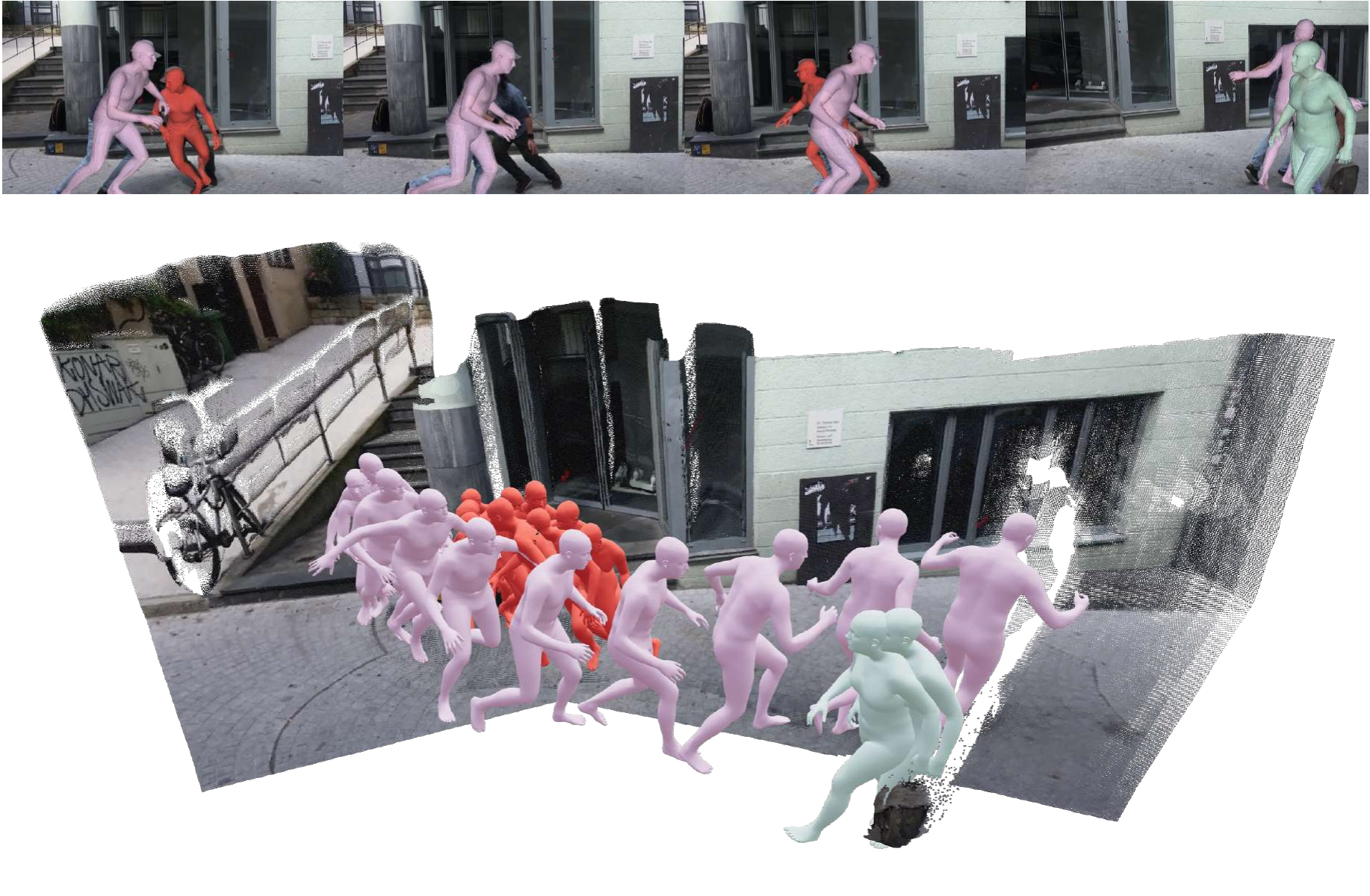}
  \caption{Temporal occlusion example.}
  \label{fig:temporal_occlusion}
\end{figure}

\vspace{2mm}\noindent\textbf{Temporal Occlusion in Monocular Setting.} 
Since our detection follows the CenterNet-based paradigm~\citep{zhou2019objects}, a person whose head is fully occluded cannot be detected. In the multi-view setting, people can be recovered with other visible views. However in a monocular setting, when the head is fully invisible, it cannot be recovered. For practicality, we track people using their refined query features and keep unmatched tracks in memory so that they can be re-matched after temporary missed detections. As shown in Fig.~\ref{fig:temporal_occlusion}, our framework consistently tracks people through occlusion without identity switches.

\subsection{Additional Experiments}
Finally, we compute runtime per component, compare with other human motion recovery methods, and analyze the effect of our scale calibration on another baseline.

\begin{wraptable}{rt}{0.35\columnwidth}
\centering
\vspace{-\baselineskip}
\caption{Runtime analysis on multi-view scene.}
\footnotesize
\setlength{\tabcolsep}{4pt}
\begin{tabular}{l|c}
\toprule
Module & Runtime $\downarrow$ \\
\midrule

Scene encoder         & 0.09s \\
Scene decoder         & 0.25s \\
Scene heads         & 0.23s \\
\midrule
Human encoder      & 0.40s \\
Human decoder       & 0.01s \\
SMPL head       & 0.02s \\
\midrule
Association         & 0.05s \\
Multi-View fusion   & 0.08s \\
\midrule
Total               & \textbf{1.13s} \\
\bottomrule
\end{tabular}
\vspace{-\baselineskip}
\label{tab:mv_inference_time}
\end{wraptable}
\vspace{2mm}\noindent\textbf{Runtime Per Component.}
To further analyze the runtime cost in Table~\ref{tab:mv_pose}, we measure the computation time of each module of our framework in Table~\ref{tab:mv_inference_time}. We select the same scene with three people and four views from EgoHumans~\citep{khirodkar2023ego}, repeat the inference five times, and report the average runtime on a single NVIDIA V100 GPU.

\vspace{2mm}\noindent\textbf{Comparison with Human Motion Recovery Methods.}
We report global motion quality with world-grounded human motion recovery methods~\citep{ye2023decoupling, shin2024wham, shen2024world, wang2024tram} on EMDB-2~\citep{kaufmann2023emdb} in Table~\ref{tab:human_only}. These methods estimate only human motion and rely on external SLAM systems such as DROID-SLAM~\citep{teed2021droid} or DPVO~\citep{teed2023deep}, whereas our framework jointly reconstructs cameras, scene, and humans in a feed-forward pass. Our method achieves competitive global motion accuracy. 
\begin{table}[h]
\centering
\scriptsize
\caption{Comparison with world-grounded human motion recovery methods on EMDB-2.}
\label{tab:human_only}
\begin{tabular}{l cc ccc}
\toprule
& \multicolumn{2}{c}{\textbf{Settings}} & \multicolumn{3}{c}{\textbf{EMDB-2}} \\
& Scene & Cam & WA-MPJPE$\downarrow$ & W-MPJPE$\downarrow$ & RTE(\%)$\downarrow$ \\
\midrule
SLAHMR & \xmark & DROID-SLAM & 326.9 & 776.1 & 10.2 \\
WHAM   & \xmark & DPVO & 135.6 & 354.8 & 6.0 \\
GVHMR  & \xmark & DPVO & 111.0 & 276.5 & 2.0 \\
TRAM   & \xmark & DROID-SLAM & \textbf{76.4} & 222.4 & \textbf{1.4} \\
\midrule
\rowcolor{gray!15}
Ours & \checkmark & Feed-Forward & 83.6 & \textbf{214.4} & \textbf{1.4} \\
\bottomrule
\end{tabular}
\end{table}

\vspace{2mm}\noindent\textbf{Scale Calibration on Human3R.}
To examine whether scale calibration transfers to other frameworks, we apply it
to Human3R by computing the ratio $r$ in Eq.~\ref{eq:calibration} from its SMPL
meshes and our 2D keypoints. We do not apply depth residual prediction, as it requires re-training. As shown in Table~\ref{tab:calib_baselines}, calibration improves global motion accuracy on EMDB-2 but slightly degrades it on RICH, as Human3R's translation is not anchored on the scene depth. This suggests that scale calibration is most effective when combined with depth anchoring.
\begin{table}[h]
\centering
\scriptsize
\caption{Scale calibration applied to Human3R without depth residual prediction.}
\label{tab:calib_baselines}
\resizebox{0.85\linewidth}{!}{
\begin{tabular}{l | ccc | ccc}
\toprule
& \multicolumn{3}{c|}{\textbf{RICH}} & \multicolumn{3}{c}{\textbf{EMDB-2}} \\
& WA-MPJPE$\downarrow$ & W-MPJPE$\downarrow$ & RTE(\%)$\downarrow$
& WA-MPJPE$\downarrow$ & W-MPJPE$\downarrow$ & RTE(\%)$\downarrow$ \\
\midrule
Human3R          & \textbf{110.0} & \textbf{184.9} & \textbf{3.3} & 112.2 & 267.9 & 2.2 \\
Human3R + Calib. & 111.2 & 190.1 & 3.5 & \textbf{106.4} & \textbf{250.4} & \textbf{2.1} \\
\bottomrule
\end{tabular}}
\end{table}

\section{Implementation Details}
\label{sec:implementation}
\subsection{Model Architecture}
\vspace{2mm}\noindent\textbf{Network Design.}
We freeze all Pi3X~\citep{wang2025pi} modules and the DINOv2~\citep{oquab2023dinov2} ViT backbone from Multi-HMR~\citep{baradel2024multi}, including the detection MLP that predicts a person-center score map and the offset MLP that localizes each human head.
For human reconstruction, we train a mask MLP to predict full-resolution person masks and a human decoder to refine the detected person tokens. The decoder consists of transformer blocks with self-attention among the person tokens, cross-attention to the image features, and a feed-forward network. Three SMPL heads decode each refined token into full body pose, shape coefficients, and a depth residual.
We also train heatmap heads implemented as two linear projections, one applied to the refined person query and the other to the human feature map, whose dot products produce per-joint heatmaps decoded using a differentiable soft-argmax.
Two additional linear heads predict the per-joint visibility and confidence logits used in Sec.~\ref{sec:bidirectional_anchoring}.
For multi-view fusion, we train a shared parameter encoder, implemented as a two-layer MLP with GELU activation, to map the concatenated 6D pose and shape parameters to the transformer width. The refine transformer consists of transformer layers with multi-head self-attention and GELU feed-forward networks. Two MLP heads decode the fused token into pose and shape residuals. Their final layers are zero-initialized so that the module initially preserves the averaged parameters.
\begin{algorithm}[t]
\caption{Multi-Person Association}
\footnotesize
\begin{algorithmic}[1]
\State \textbf{1. Per-View Tracking}
\For{each view $v = 1, \dots, V$}
    \State Initialize human track IDs and the track memory with all detections at $t=1$
    \For{frame $t = 2, \dots, T$}
        \State Compute pairwise L2 distance matrix $\mathcal{D}$ between tokens in the track memory and detections at $t$
        \State Perform one-to-one matching via Sinkhorn optimal transport
        \State Assign new human track IDs to unmatched detections
        \State Update the track memory with matched tokens; keep unmatched tracks for re-matching
    \EndFor
    \For{each human tracklet $m$}
        \State Transform 3D joints of tracklet $m$ into world coordinates using camera extrinsics
        \For{each observed frame $t$ of tracklet $m$ with previous observed frame $t'$}
            \State Compute mean joint displacement
            $\Delta = \frac{1}{J}\sum_j \|\mathcal{J}^{m}_{t,j}-\mathcal{J}^{m}_{t',j}\|_2$
            \If{$\Delta > \gamma_{\text{outlier}}$}
                \State Remove frame $t$ from human tracklet $m$
            \EndIf
        \EndFor
    \EndFor
\EndFor

\State

\State \textbf{2. Cross-View Association}
\State Compute mean camera center for each view
\State Construct view connectivity graph using k-NN
\For{each connected view pair $(v_i,v_j)$}
    \For{each human tracklet pair $(a,b)$}
        \State Compute matching cost $\mathcal{C}(a,b)$ using Eq.~\ref{eq:matching_cost}
    \EndFor
    \State Solve one-to-one assignment via Hungarian algorithm
    \State Retain matches with $\mathcal{C} < \gamma_{\text{reid}}$
\EndFor
\State Merge retained matches across all view pairs via union-find
\State Assign one global human ID per connected component

\end{algorithmic}
\label{alg:mp_assot_algo}
\end{algorithm}

\vspace{2mm}\noindent\textbf{Association Algorithm.}
Next, we describe the multi-person association algorithm in Alg.~\ref{alg:mp_assot_algo}. Given the refined human queries \(\hat{Q}\) and SMPL 3D joints \(\mathcal{J}^{\text{3D}}\) from Sec.~\ref{sec:model_architecture}, we first track humans within each view by matching the queries of the current detections to a track memory containing the stored query of each observed track. Matching is performed using pairwise L2 distances and Sinkhorn optimal transport. Matched tracks update their stored queries, while unmatched tracks remain in memory so that they can be re-matched after temporary misses. We then filter outliers based on the average displacement of the world-space 3D joints between consecutive observations of each tracklet. After per-view tracking, we associate human tracklets across views by applying the Hungarian algorithm to the matching cost defined in Eq.~\ref{eq:matching_cost}. Finally, the matched tracklets are merged using union-find, and global human identities are assigned.

\subsection{Training Setup}
We train our framework in three stages on 4 NVIDIA A6000 GPUs. Each stage takes approximately one day, resulting in a total training time of about three days. Throughout training, we set the maximum number of people per view to 10.

\vspace{2mm}\noindent\textbf{Training Schedule.}
In Stage 1, we freeze the scene encoder, scene decoder, and scene output heads, as well as the human encoder and the detection and offset MLPs. We train the human decoder, SMPL head, mask MLP, heatmap heads, and visibility and confidence heads for 20 epochs. Scale calibration is disabled for the first 10 epochs to focus on learning human shape, pose, and translation, and is enabled for the remaining 10 epochs. For each training sample, the input sequence length is randomly sampled between 2 and 24 frames. We use the AdamW optimizer with a OneCycleLR scheduler and an initial learning rate of \(5\times10^{-5}\). Ground-truth depth maps, camera parameters, and head keypoint locations are provided as input condition during this stage. The multi-view fusion module is disabled.

In Stage 2, we additionally freeze the human decoder, SMPL head, and mask MLP, and train only the heatmap, visibility, and confidence heads for 20 epochs. The model takes a single image as input, with only the head keypoint locations provided, while the camera parameters are predicted by the scene branch. We use the AdamW optimizer with a OneCycleLR scheduler and an initial learning rate of \(1\times10^{-4}\). The multi-view fusion module remains disabled.

In Stage 3, we freeze all modules except the multi-view fusion module and train it for 20 epochs. Specifically, we train the shared parameter encoder, refine transformer, and pose and shape residual heads. Each training sample contains a single time step with 2 to 8 views, and ground-truth person identities across views are provided. Ground-truth head keypoint locations and camera intrinsics are provided in every batch, whereas ground-truth camera extrinsics are provided in a randomly selected 50\% of the batches. We use the AdamW optimizer with a OneCycleLR scheduler, starting with a learning rate of $1\times10^{-4}$. To improve robustness to unreliable views, we augment the lower body of a randomly selected person in one randomly selected view by pasting Pascal VOC~\citep{everingham2010pascal} objects to simulate occlusion and by applying a random crop to simulate truncation. The two augmentations are applied independently, and the probability of each increases linearly from 0 to 0.5 over the first 5 epochs.
At inference, all camera parameters are predicted by the scene branch without ground-truth extrinsics.

\vspace{2mm}\noindent\textbf{Dataset Configuration.}
Across all stages, we use dynamic-resolution training with aspect ratios ranging from \(0.5\) to \(2.0\). We augment the images using random cropping, focal-length jitter, color jitter, and blurring. Stage~1 is trained only on BEDLAM~\citep{black2023bedlam}, a synthetic dataset that provides SMPL-X parameters, depth maps, and camera parameters for every frame. We compute per-joint occlusion labels for the seven detector joints by depth-testing the projected SMPL-X joints against the rendered depth maps. To improve performance on in-the-wild scenarios, Stage~2 is trained on single images from BEDLAM, 3DPW~\citep{von2018recovering}, MPII~\citep{andriluka20142d}, and MSCOCO~\citep{lin2014microsoft}, sampled at a ratio of \(1{:}1{:}1{:}3\). For the three in-the-wild datasets, we use SMPL-X pseudo-ground truth from NeuralAnnot~\citep{moon2022neuralannot} and precompute the 2D projections of the joints and vertices to supervise the 2D joint loss. Occlusion labels are obtained from OpenPose~\citep{cao2019openpose} confidence scores for 3DPW, keypoint visibility flags for MSCOCO, and visibility annotations for MPII. Joints without an available label are excluded from the confidence loss. For Stage~3, we train the multi-view fusion module on MammaSyn~\citep{cuevas2026mamma}, a synthetic dataset containing approximately 895 two-person interaction sequences rendered from eight synchronized cameras, with world-space SMPL-X parameters and cross-view-consistent person identities. Per-joint occlusion labels are derived from the per-view vertex visibility provided by the dataset.

\vspace{2mm}\noindent\textbf{Loss Design.}
In Stage~1, we use geometric, parametric, and detection losses. The geometric losses \(\mathcal{L}_{\text{v3d}}\) and \(\mathcal{L}_{\text{j3d}}\) supervise the predicted SMPL vertices and 3D joints in both camera and pelvis-relative coordinates, after aligning the predicted translation and scale with the ground truth using the scale factor \(s_\text{opt}\) from the ROE solver~\citep{wang2025moge}. Their 2D projections are supervised by \(\mathcal{L}_{\text{v2d}}\) and \(\mathcal{L}_{\text{j2d}}\). The parametric losses supervise the SMPL pose \(\theta^\text{full}\), shape \(\beta\), and translation \(\tau\), and since scale calibration depends on body size, an L1 loss \(\mathcal{L}_{\text{height}}\) additionally supervises the rest-pose body height derived from \(\beta\). The detection losses consist of a BCE mask loss \(\mathcal{L}_{\text{mask}}\), an image-size-normalized L1 loss \(\mathcal{L}_{\text{loc}}\) on in-frame detector joints, and BCE losses \(\mathcal{L}_{\text{vis}}\) and \(\mathcal{L}_{\text{conf}}\) on the visibility and confidence logits, where \(\mathcal{L}_{\text{conf}}\) is applied only to joints with occlusion labels:
\begin{equation}
\begin{aligned}
\mathcal{L}_{\text{stage1}} =\;&
\lambda_\text{v3d}\mathcal{L}_{\text{v3d}} +
\lambda_\text{j3d}\mathcal{L}_{\text{j3d}} +
\lambda_\text{v2d}\mathcal{L}_{\text{v2d}} +
\lambda_\text{j2d}\mathcal{L}_{\text{j2d}} +
\lambda_{\theta}\mathcal{L}_{\theta^\text{full}} +
\lambda_\beta\mathcal{L}_{\beta} +
\lambda_\tau\mathcal{L}_{\tau} \\
&+
\lambda_\text{height}\mathcal{L}_{\text{height}} +
\lambda_\text{mask}\mathcal{L}_{\text{mask}} +
\lambda_\text{loc}\mathcal{L}_{\text{loc}} +
\lambda_\text{vis}\mathcal{L}_{\text{vis}} +
\lambda_\text{conf}\mathcal{L}_{\text{conf}},
\end{aligned}
\label{eq:stage1_loss}
\end{equation}
where \(\lambda_\text{v3d}=\lambda_\text{j3d}=\lambda_\tau=5.0\) and all other weights are \(1.0\).

In Stage~2, since the in-the-wild datasets lack ground-truth cameras, the 2D reprojection losses are computed with the predicted intrinsics and without \(s_\text{opt}\), and are combined with the joint detection losses:
\begin{equation}
\mathcal{L}_{\text{stage2}} =
\lambda_\text{v2d}\mathcal{L}_{\text{v2d}} +
\lambda_\text{j2d}\mathcal{L}_{\text{j2d}} +
\lambda_\text{loc}\mathcal{L}_{\text{loc}} +
\lambda_\text{vis}\mathcal{L}_{\text{vis}} +
\lambda_\text{conf}\mathcal{L}_{\text{conf}},
\label{eq:stage2_loss}
\end{equation}
with all weights set to \(1.0\).

In Stage~3, we supervise the fused SMPL parameters from the multi-view fusion module. The fused root rotation and translation are transformed into each input camera frame and passed through the SMPL-X layer. We apply \(\mathcal{L}_{\theta^\text{full}}\), \(\mathcal{L}_\beta\), \(\mathcal{L}_{\text{j2d}}\), and \(\mathcal{L}_{\text{v2d}}\), and since the fused body is anchored at the triangulated head, \(\mathcal{L}_{\text{j3d}}\) and \(\mathcal{L}_{\text{v3d}}\) are computed in head-relative coordinates. An L2 penalty \(\mathcal{L}_\Delta\) regularizes the pose and shape residuals:
\begin{equation}
\begin{aligned}
\mathcal{L}_{\text{stage3}} =\;&
\lambda_\text{v3d}\mathcal{L}_{\text{v3d}} +
\lambda_\text{j3d}\mathcal{L}_{\text{j3d}} +
\lambda_\text{v2d}\mathcal{L}_{\text{v2d}} +
\lambda_\text{j2d}\mathcal{L}_{\text{j2d}} \\
&+
\lambda_{\theta}\mathcal{L}_{\theta^\text{full}} +
\lambda_\beta\mathcal{L}_{\beta} +
\lambda_\Delta\mathcal{L}_{\Delta},
\end{aligned}
\label{eq:stage3_loss}
\end{equation}
where \(\lambda_\text{v3d}=\lambda_\text{j3d}=5.0\), \(\lambda_\Delta=0.01\), and all other weights are \(1.0\).

\subsection{Evaluation Protocols}
For global human motion estimation, we evaluate on EMDB-2~\citep{kaufmann2023emdb}, which captures monocular dynamic-camera videos, and RICH~\citep{huang2022capturing}, which provides fixed multi-view cameras.
We measure global motion accuracy in the world coordinate system using three metrics. World-Aligned MPJPE (WA-MPJPE) aligns the predicted joints of each 100-frame segment to the ground truth using a Sim(3) transformation, while World MPJPE (W-MPJPE) estimates this alignment from only the first two frames of each segment. Both metrics are reported in millimeters. We also report the Root Translation Error (RTE, \%), which measures the root translation error after SE(3) alignment over the entire sequence, normalized by the total displacement to account for varying motion magnitudes. For multi-view human pose estimation, we follow prior single-frame protocols on EgoHumans~\citep{khirodkar2023ego} and EgoExo4D~\citep{grauman2024ego}. We report W-MPJPE$^\dagger$ after camera-only SE(3) alignment, GA-MPJPE after jointly aligning all people with a single Sim(3) transformation, and PA-MPJPE after person-wise Procrustes alignment. All three metrics are reported in meters.

\section{Additional Results}
\label{sec:results}
\subsection{Qualitative Results}
\begin{figure}[!t]
  \centering
  \includegraphics[width=\linewidth]{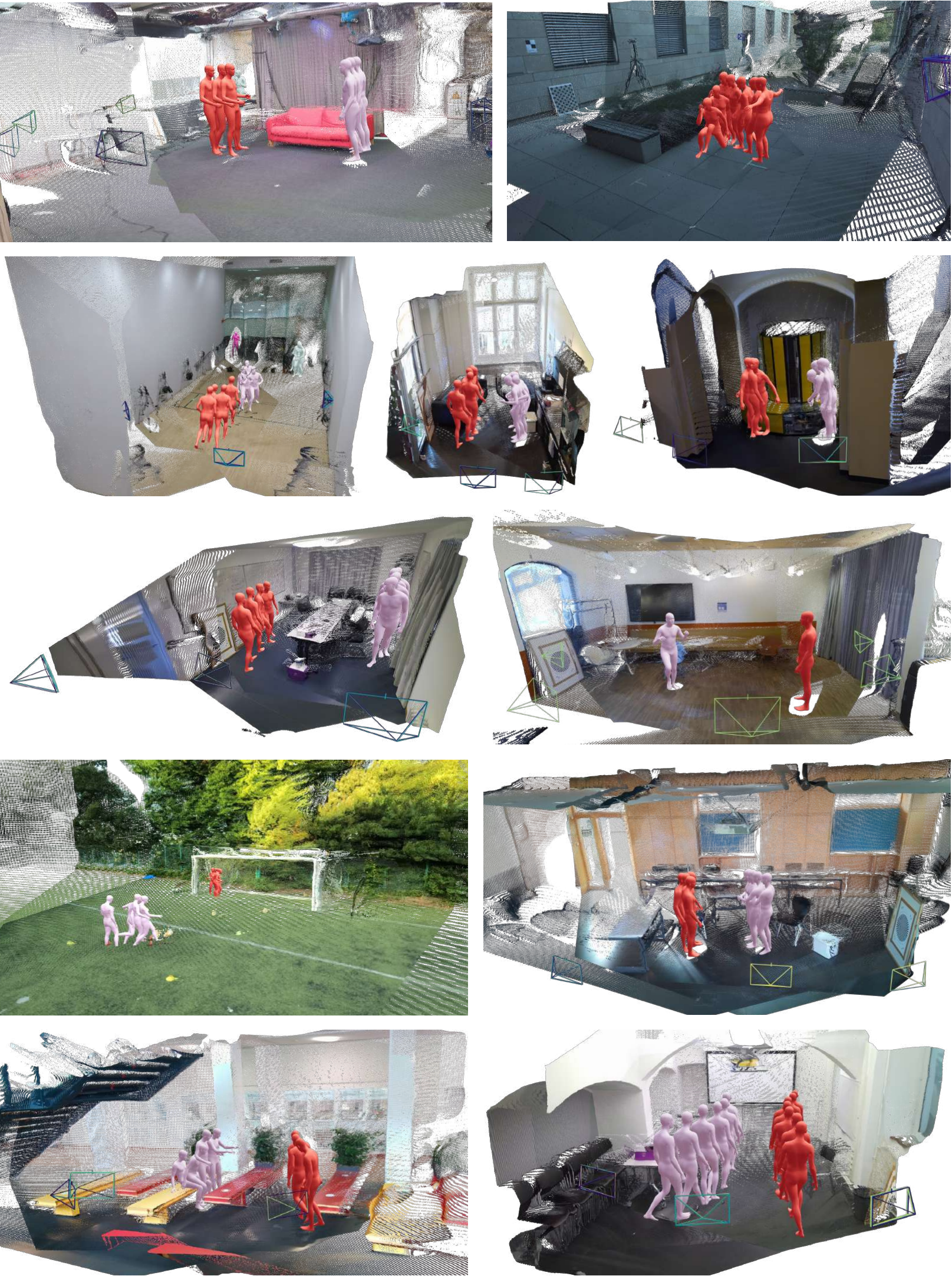}
  \caption{Qualitative results in multi-view settings.}
  \label{fig:supp_multi}
  \vspace{-5mm}
\end{figure}

\begin{figure}[!t]
  \centering
  \includegraphics[width=\linewidth]{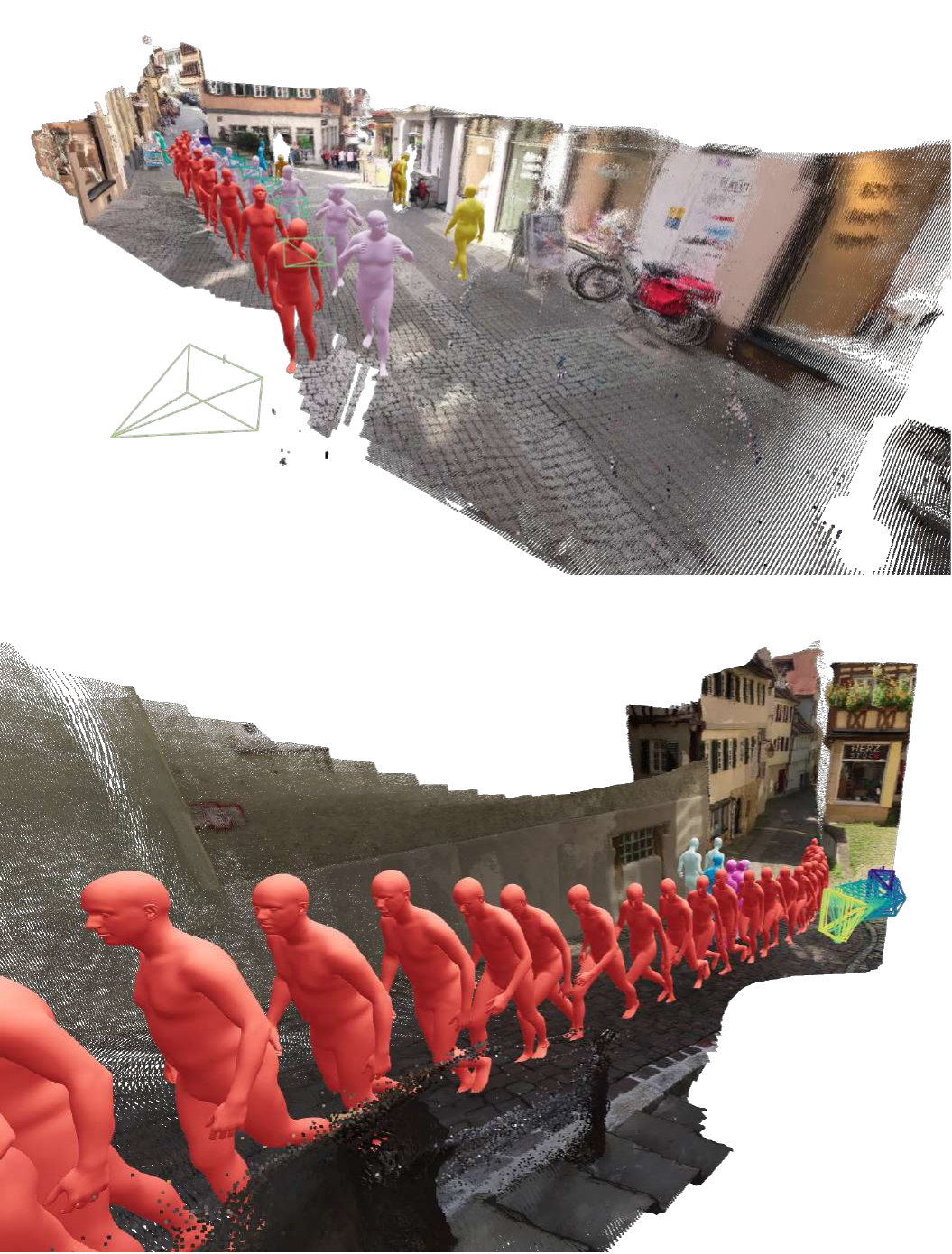}
  \caption{Qualitative results in monocular settings.}
  \label{fig:supp_mono}
  \vspace{-2mm}
\end{figure}

\begin{figure}[!t]
  \centering
  \includegraphics[width=\linewidth]{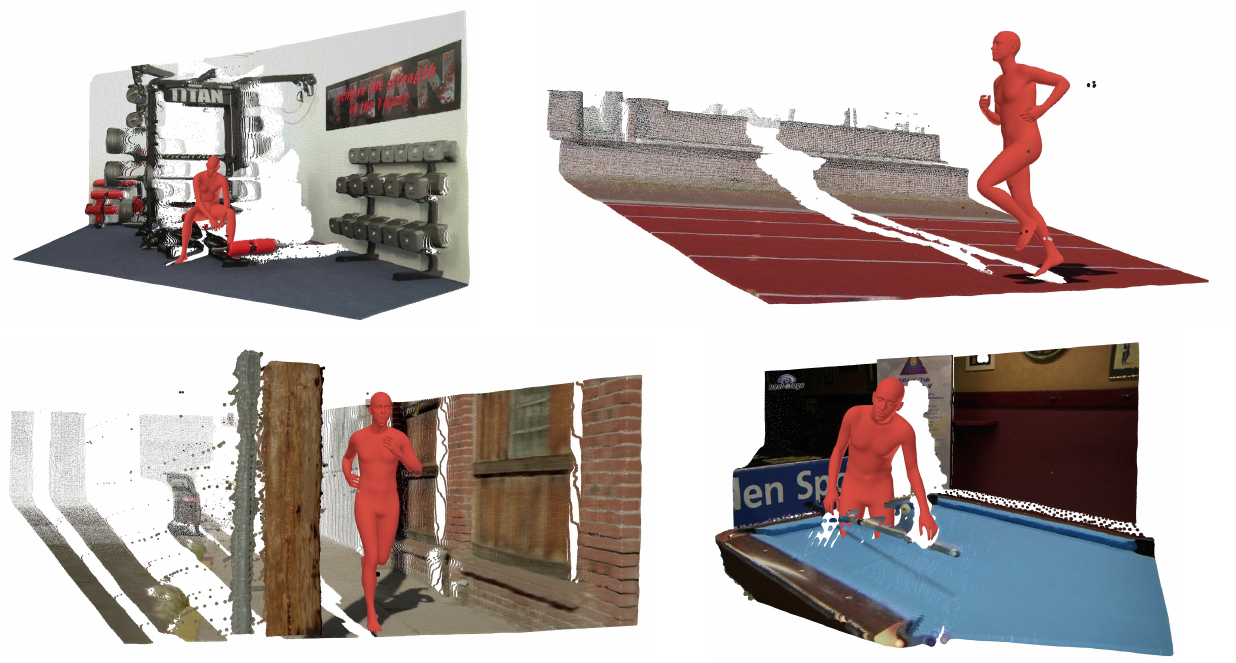}
  \caption{Qualitative results on single images.}
  \label{fig:supp_single}
  \vspace{-2mm}
\end{figure}

In this section, we present additional qualitative results under various scenarios. As illustrated in Fig.~\ref{fig:supp_multi}, our framework reconstructs humans and the scene in diverse multi-view settings, including EgoBody~\citep{zhang2022egobody}. As shown in Fig.~\ref{fig:supp_mono} and Fig.~\ref{fig:supp_single}, our framework can also handle monocular and single-image scenarios. \ours~is capable of reconstructing both humans and the scene under both multi-view and monocular setups in a single feed-forward pass, which shows its practicability.

\subsection{Failure Cases}
\begin{figure}[!t]
  \centering
  \includegraphics[width=\linewidth]{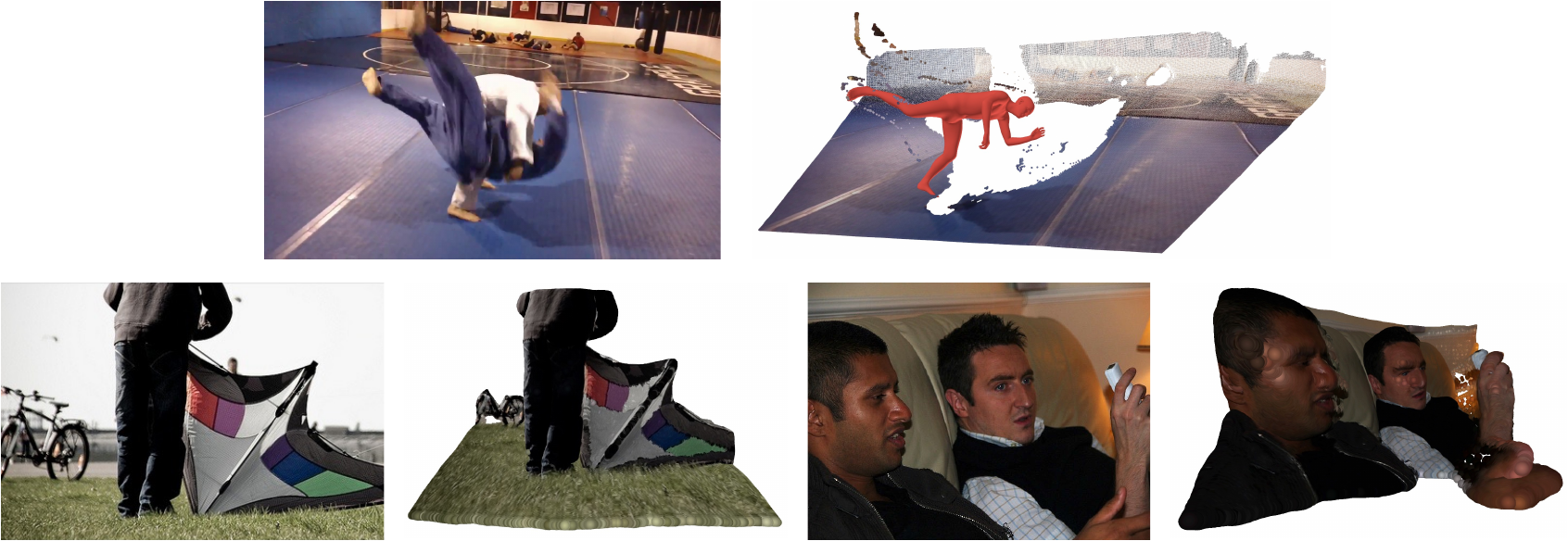}
  \caption{Failure cases.}
  \label{fig:supp_fail}
  \vspace{-2mm}
\end{figure}

In this section, we present failure cases where our framework fails to reconstruct humans and the scene. As shown in Fig.~\ref{fig:supp_fail}, our framework exhibits limitations when handling complex poses and close human interactions. It also struggles when the head is completely invisible in the image. Furthermore, the model faces challenges in extreme zoom-in scenarios where the head occupies most of the image. We leave these limitations for future work.

\label{sec:limitations}
\clearpage

\end{document}